\documentclass[11pt]{article}

\usepackage[preprint]{acl}

\usepackage{times}
\usepackage{latexsym}

\usepackage[T1]{fontenc}

\usepackage[utf8]{inputenc}

\usepackage{microtype}

\usepackage{inconsolata}

\usepackage{graphicx}
\usepackage{amsmath}
\usepackage{amssymb}

\usepackage{booktabs}
\usepackage{multirow}
\usepackage{tabularx}

\usepackage{pifont}
\newcommand{\cmark}{\ding{51}}
\newcommand{\xmark}{\ding{55}}

\usepackage{listings}
\usepackage{xcolor}
\definecolor{codebg}{RGB}{240,240,240}

\usepackage{placeins}
\title{Dependency Reasoning for Aspect Sentiment Quadruple Prediction}

\author{
  \textbf{Shima Esfandyari}$^{1}$,
  \textbf{Seywan Soleymanbaigi}$^{1}$,\\
  \textbf{Fardin Akhlaghian Tab}$^{1}$,
  \textbf{Fatemeh Daneshfar}$^{1}$\thanks{Corresponding author.},\\
  \textbf{Mourad Oussalah}$^{2}$\\[6pt]
  $^{1}$Department of Computer Engineering, University of Kurdistan, Sanandaj, Iran\\
  $^{2}$University of Oulu, Finland\\
  \texttt{\{shima.esfandyari,s.soleymanbaigi,f.akhlaghian,f.daneshfar\}@uok.ac.ir}\\
  \texttt{Mourad.Oussalah@oulu.fi}
}

\begin{document}
\maketitle
\begin{abstract}
Aspect Sentiment Quadruple Prediction (ASQP) is fundamentally a relational reasoning problem: identifying aspect--opinion--category--sentiment quadruples requires modeling structured dependencies, not merely sequential co-occurrence. Existing LLM-based approaches incorporate syntactic information indirectly, via retrieval-based demonstration selection or multi-stage training, rather than as unified, single-pass generation-time context. We present DRIFT, which linearizes dependency parse trees and injects them directly into LLM prompts as explicit relational context, enabling the model to reason over aspect--opinion dependencies rather than infer them from token proximity alone. DRIFT further integrates retrieval-augmented in-context learning and a syntax-aware LLM-based adjudication mechanism that complements exact-match evaluation under span-boundary variation. On SemEval Rest15, Rest16, and ACOS, DRIFT achieves the strongest performance among compared generative ASQP methods, reaching 54.47\% and 64.10\% F1 on Rest15 and Rest16, with consistent gains across short- and medium-distance aspect--opinion configurations; adjudication further shows exact-match metrics systematically underestimate semantically correct predictions\footnote{\href{https://github.com/repanonymous/RepoDRIFT}{\textnormal{https://github.com/repanonymous/RepoDRIFT}}}.
\end{abstract}

\section{Introduction} 
\begin{figure}[t]
\centering
\includegraphics[width=0.98\columnwidth]{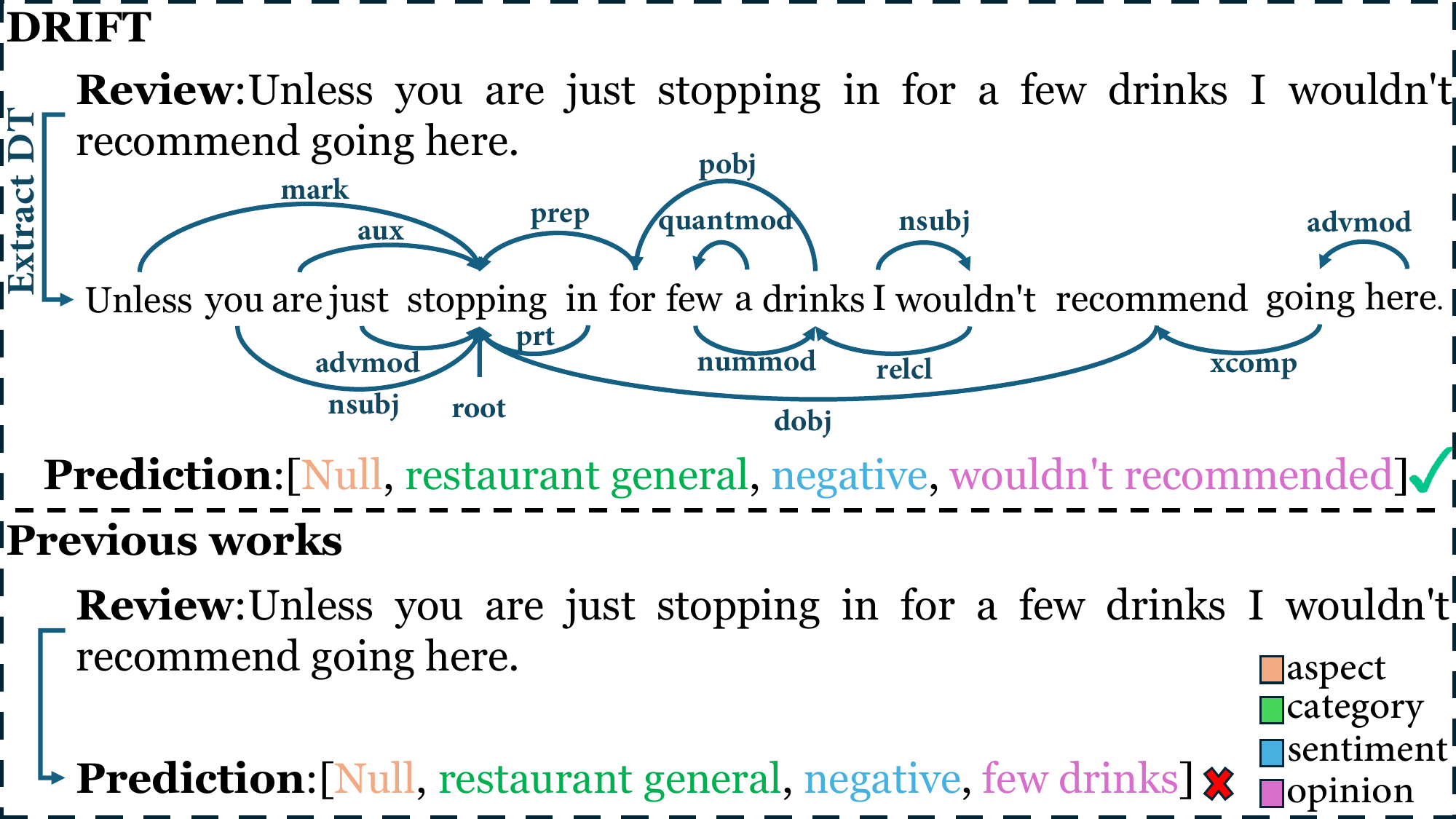}
\caption{Comparison of surface-sequential reasoning in prior ASQP approaches and dependency-aware reasoning in DRIFT for aspect--opinion alignment. Note: the example sentence is quoted verbatim from the Rest15 dataset.}
\label{fig:motivation}
\end{figure}

Aspect Sentiment Quadruple Prediction (ASQP) extracts four elements from reviews: aspect term, aspect category, sentiment polarity, and opinion term \citep{Zhang2021ASQP}. For example, ``Can't wait to go back'' yields the quadruple [Null, restaurant general, positive, go back]. Unlike pipeline-based approaches that process these elements independently \citep{Wan2020TASD}, ASQP requires joint modeling of compositional sentiment relations across all components \citep{Mao2021JointASQP}. This makes the task particularly challenging in sentences containing implicit aspects, long-range aspect--opinion dependencies, or multiple overlapping sentiment expressions \citep{Cai2021ACOS, Zhang2024SDGCN, Liang2022BSGCN}. Existing ASQP methods predominantly rely on \textit{surface-sequential reasoning}, where sentiment associations are inferred mainly from token proximity rather than explicit relational structure \citep{zhang-etal-2021-towards-generative, hu-etal-2022-improving-aspect, scaria-etal-2024-instructabsa}, as illustrated in Figure~\ref{fig:motivation}.

The core difficulty of ASQP is not lexical but relational: a sentence like
``Unless you are just stopping in for a few drinks I wouldn't recommend going here''
requires identifying the implicit aspect category \textit{restaurant general} and
its relation with the negative opinion expression \textit{wouldn't recommend},
while avoiding misleading surface cues such as \textit{few drinks}. This example
highlights that ASQP requires modeling aspect--opinion dependencies beyond lexical
proximity. Such aspect--opinion relations may involve long dependency paths,
cross coordination boundaries, or implicit aspects without surface mentions,
making them difficult to infer from sequential prompts that expose structural
relations only implicitly.

Recent advances in Large Language Models (LLMs) have demonstrated strong capabilities in structured reasoning, instruction following, and information extraction \citep{Wei2022ChainOfThought, NEURIPS2020_1457c0d6}. These capabilities have enabled instruction-tuned ASQP systems \citep{scaria-etal-2024-instructabsa} and retrieval-augmented methods \citep{zheng-etal-2024-instruction} that improve in-context learning (ICL) by selecting semantically relevant demonstrations. More recently, SimRP \citep{Jian_Chen_Li_Wang_Zeng_Yao_An_Wu_2025} incorporates syntactic similarity into retrieval selection. Despite these advances, existing methods primarily incorporate syntax either through retrieval-based demonstration selection or through multi-stage training procedures that decompose the task into intermediate structural sub-steps, rather than through a single unified generation-time prompt. Modern LLMs already encode substantial implicit syntactic knowledge. The challenge is therefore not syntax itself, but explicitly exposing dependency relations during generation for relational reasoning. Dependency trees explicitly encode aspect--opinion relations, complementing LLMs' implicit syntactic knowledge.

To address this gap, we propose DRIFT, a structure-grounded framework for explicit relational reasoning in ASQP. DRIFT linearizes dependency parse trees and injects them directly into LLM prompts as structured dependency context. Unlike prior work that uses syntax only for retrieval or encoder-level graph modeling, DRIFT exposes dependency relations directly within the autoregressive reasoning context, enabling dependency-guided reasoning over syntactic paths rather than token proximity.

We further extend dependency-aware reasoning to evaluation. Standard ASQP evaluation relies primarily on exact matching, which is highly sensitive to minor span-boundary variation \citep{Zhang2021ASQP, Hua2024FTSOBP}: predictions such as ``delicious'' and ``delicious food'' may express identical sentiment semantics while being penalized as incorrect. Such discrepancies reflect natural annotation flexibility rather than genuine errors \citep{yang-etal-2025-single, zhang-etal-2023-span}. We address this by introducing a syntax-aware LLM-as-a-Judge adjudication mechanism \citep{Zheng2023JudgingLLM, Liu2023GEVAL} that assesses semantic equivalence between predictions and references conditioned on the dependency structure of the input sentence, providing a more reliable complement to exact-match metrics. Our contributions are summarized as follows: 
\begin{itemize}

\item We introduce a dependency-guided prompting framework for ASQP that explicitly exposes dependency relations without architectural changes.

\item We show that dependency-guided prompting and retrieval augmentation provide complementary benefits across PEFT and ICL.

\item We introduce a syntax-aware LLM-as-a-Judge that complements exact-match evaluation by recognizing semantically equivalent predictions.

\item Experiments on Rest15, Rest16, and ACOS show the best performance among the compared generative ASQP methods, with strong cross-domain generalization.

\end{itemize}


\section{Related Work}
\label{sec:related-work}
Existing ASQP approaches can be broadly grouped into three paradigms: pipeline-based methods \citep{Wan2020TASD}, generative approaches that jointly model all quadruple elements \citep{zhang-etal-2021-towards-generative, hu-etal-2022-improving-aspect, gou-etal-2023-mvp, wang-etal-2023-generative}, and recent LLM-based frameworks leveraging instruction tuning or in-context learning \citep{scaria-etal-2024-instructabsa, zheng-etal-2024-instruction}. While these methods improve unified quadruple extraction, most still rely primarily on implicit sequential reasoning over surface-form text representations.

To better capture aspect--opinion interactions, prior work has incorporated structural inductive biases through graph-based encoders \citep{Zhang2019GCN, HuangCarley2020Attention}, implicit variable modeling \citep{nie2024modeling}, and uncertainty-aware or contrastive objectives \citep{hu-etal-2023-uncertainty, Li2024OpinionTreeguidedCL}. These approaches typically encode syntactic information internally within neural architectures, but do not explicitly expose dependency relations as reasoning guidance during generation. More recently, SimRP \citep{Jian_Chen_Li_Wang_Zeng_Yao_An_Wu_2025} incorporates syntactic similarity into retrieval-based demonstration selection for in-context learning, G2C \citep{he2026g2c} addresses exposure bias through a generate-then-correct paradigm, and S$^2$IT \citep{chen-etal-2025-s2it} serializes dependency relations into natural-language descriptions supplied to an LLM through a multi-stage training pipeline consisting of global syntax-guided extraction, local syntax-guided classification, and fine-grained structural tuning. Unlike SimRP and G2C, S$^2$IT does expose syntactic structure directly to the LLM, closely relating it to DRIFT.

DRIFT differs from S$^2$IT in how dependency structure is operationalized, not in whether it is exposed to the LLM: DRIFT supplies the complete linearized dependency tree within a single unified prompt for one-pass quadruple generation, without task decomposition or dedicated structural-tuning stages. The same formulation applies identically to parameter-efficient fine-tuning and to black-box in-context learning, and extends to dependency-conditioned semantic adjudication (Section~\ref{sec:adjudication}). This yields a simple, model-agnostic way to use explicit dependency context, complementary to S$^2$IT's stepwise, training-oriented formulation. Detailed discussion of related work is provided in Appendix~\ref{app:related-work}.

\section{Method}

\begin{figure*}[!t]
    \centering
    \includegraphics[width=0.81\textwidth]{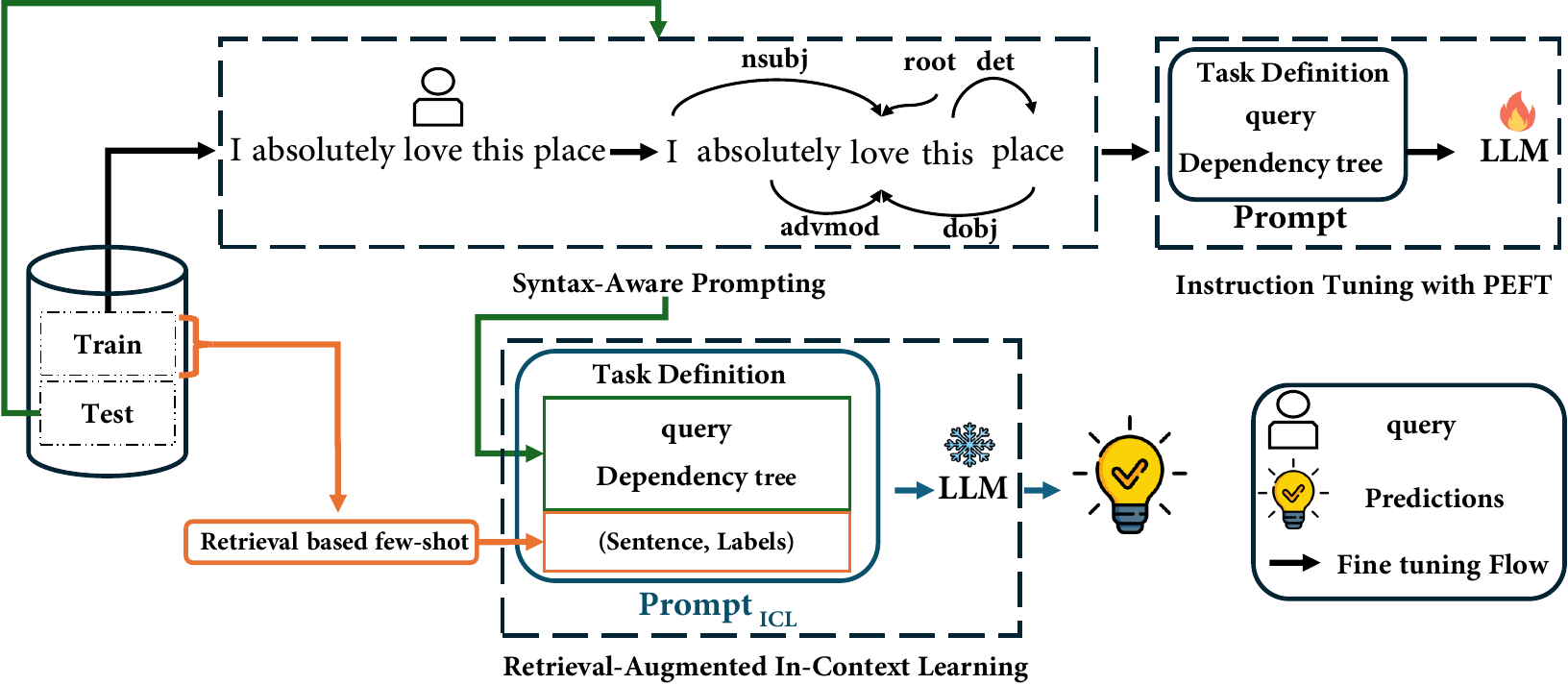}
    \vspace{-1 pt}
    \caption{
        Overview of the DRIFT framework with dependency-guided prompting and retrieval augmentation.
    }
    \label{fig:method}
\end{figure*}

We propose DRIFT, a structure-grounded framework for ASQP that integrates syntax-aware prompting, Parameter-Efficient Fine-Tuning (PEFT), and retrieval-augmented in-context learning (Figure~\ref{fig:method}).

The central insight behind DRIFT is that ASQP is not a sequence labeling problem but a relational reasoning problem. A sequential prompt presents the model with a string of tokens and implicitly asks it to infer aspect--opinion dependencies from word order and co-occurrence. A dependency-augmented prompt instead makes these relations symbolically explicit: the model receives a structured description of which tokens are connected, through which grammatical relation, and in which direction. This transforms the model's reasoning task from \textit{relation inference} (hard) to \textit{relation instantiation} (easier), reducing the effective search space over possible aspect--opinion pairings and making long-range dependencies tractable without modifying model parameters or architecture.

At inference time, DRIFT constructs prompts combining the task definition, input sentence, and linearized dependency tree. Semantically similar demonstrations are retrieved to provide in-context examples. An LLM-based adjudicator then evaluates semantic equivalence between predictions and reference quadruples, addressing annotation inconsistencies.

A key property of DRIFT is its model-agnostic and domain-flexible design. Operating entirely through prompt construction and syntactic augmentation, it applies to encoder--decoder, decoder-only, and black-box LLMs without architectural modifications. Our experiments span restaurant reviews (SemEval Rest15, Rest16, Restaurant-ACOS) and laptop reviews (Laptop-ACOS), confirming cross-domain generalization.

\subsection{Problem Formulation}

Given an input sentence \(x = (w_1, \dots, w_n)\), the goal of ASQP is to extract a set of aspect–sentiment quadruples:
\begin{equation}
Y(x) = \{\, q_i \,\}_{i=1}^{m}, \qquad 
q_i = (a_i, c_i, s_i, o_i)
\end{equation}
where \(a_i\) denotes the \emph{aspect term}, \(o_i\) the \emph{opinion term}, \(c_i \in \mathcal{C}\) the \emph{aspect category} selected from a predefined category set \(\mathcal{C}\), and \(s_i \in \{\textit{Positive}, \textit{Negative}, \textit{Neutral}\}\) the \emph{sentiment polarity}. The variable \(m\) is the number of aspect–opinion quadruples identified in the sentence \(x\).


\subsection{Syntax-Aware Prompting}
\label{sec:syntax-aware-prompting}

Given an input sentence \(x\), we obtain its dependency parse \(D(x)\) using spaCy.\footnote{\href{https://spacy.io/}{\textnormal{https://spacy.io/}}} We then construct a syntax-aware prompt:

\begin{equation}
Prompt(x) = [\, task;\, x;\, D(x) \,].
\end{equation}

The dependency tree is linearized into ordered head--relation--dependent triples:
\[
(\text{head}, \text{relation}, \text{dependent}),
\]
where directionality is explicitly preserved as:
\[
\text{head} \rightarrow \text{relation} \rightarrow \text{dependent}.
\]

A deterministic head-first traversal is used to preserve relational consistency while minimizing prompt length. Final dependency sequences are represented as semicolon-separated triples (Appendix~\ref{app:linearization}).

This formulation serves a specific reasoning function: it converts implicit syntactic relations which the model must infer from token co-occurrence into explicit symbolic constraints available at each decoding step. When the model generates an opinion term, the dependency tree makes it immediately verifiable which tokens are syntactically connected to that term, reducing the model's relational search space and improving precision in aspect-opinion alignment. Crucially, this changes the reasoning \textit{process} rather than merely the input representation.

This design is motivated by ASQP's structure: unlike single-label ABSA subtasks, ASQP requires linking each opinion to the correct aspect/category among several candidates, a decision proximity-based reasoning gets wrong when the true predicate–argument relation is linearly distant or competes with a nearer, syntactically unrelated phrase. For example, in ``...this spot delivers,'' \textit{delivers} is linearly closer to \textit{modern Japanese} than to \textit{spot}, yet the dependency path correctly identifies \textit{spot} as its syntactic subject (Appendix~\ref{app:qualitative}). Section~\ref{sec:dep-complexity} tests this directly as a function of dependency distance.

\subsection{Instruction Tuning with PEFT}

For efficient adaptation, a causal language model is fine-tuned using QLoRA on instruction-formatted samples. Each training instance is a pair
\(\big(Prompt(x),~ y\big)\), where \(y = \{(a_i, c_i, s_i, o_i)\}_i\) is the target quadruple set. The loss is computed over target tokens only:
\begin{equation}
\mathcal{L}_{\text{SFT}}
= -\sum_{t=1}^{T}
\log p_\theta\!\bigl(y_t \mid y_{<t}, Prompt(x)\bigr),
\end{equation}
where \(p_\theta(\cdot \mid y_{<t}, Prompt(x))\) is parameterized by trainable adapter parameters \(\theta\), while the base model remains frozen.

We apply Low-Rank Adaptation (LoRA) \citep{lora2022hu} to selected transformer layers. Each weight matrix \(W \in \mathbb{R}^{d \times k}\) is decomposed as:
\begin{equation}
W' = W + \Delta W, \quad \Delta W = BA,
\end{equation}
where \(A \in \mathbb{R}^{r \times k}\) and \(B \in \mathbb{R}^{d \times r}\) are trainable low-rank matrices (\(r \ll \min(d, k)\)), while \(W\) remains frozen. QLoRA \citep{NEURIPS2023_1feb8787} further reduces memory usage by loading the base model in 4-bit NF4 quantization, keeping LoRA adapters in bfloat16 precision.

\subsection{Retrieval-Augmented ICL}

During inference, semantically similar demonstration pairs
\(S(x)=\{(x_j, y_j)\}_{j=1}^{k}\) are retrieved using a frozen Sentence Transformer encoder. The final ICL prompt is constructed as:
\begin{equation}
Prompt_{\mathrm{ICL}}(x)=\bigl[\, task;~ S(x);~ x;~ D(x) \,\bigr].
\end{equation}

This combines semantic retrieval with syntax-aware dependency sequences. Semantic similarity selects demonstrations with matching opinion expressions, while dependency trees expose structural relations for accurate aspect–opinion reasoning. The separation of semantic retrieval and syntactic guidance ensures both relevance and structural accuracy during one-pass quadruple generation (Appendix~\ref{app:prompts}).

\subsection{LLM-Based Adjudication}
\label{sec:adjudication}
To address span-boundary inconsistencies in ASQP evaluation, we introduce a blind semantic adjudicator \(J\) that compares two quadruple sets \(U\) and \(V\) conditioned on the task description, input sentence \(x\), and dependency structure \(D(x)\):
\begin{equation}
(d, r) = J(U,V \mid [task, x, D(x)]),
\end{equation}

where \(d \in \{0,1\}\) denotes the adjudication decision and \(r\) represents textual reasoning. The judge is not informed which set corresponds to the prediction or reference, reducing positional bias. Semantically equivalent outputs (e.g., ``delicious'' vs.\ ``delicious food'') can therefore be treated as correct despite boundary variation.

The adjudication framework is model-agnostic and can be applied to outputs from both fine-tuned and prompting-based systems. Additional experiments with multiple judge models demonstrate that evaluation trends remain stable across adjudicators (Table~\ref{tab:judge_models}).
The semantic adjudication component is evaluated independently during evaluation and is not illustrated in Figure~\ref{fig:method}.

\section{Experiments}
The detailed descriptions of the datasets, evaluation metrics, and implementation setup are provided in Appendix~\ref{app:set-up}.

\subsection{Main Results}
\label{sec:main-results}
Following the ablation findings discussed in Section~\ref{sec:ablation}, the primary DRIFT results use the PEFT configuration with dependency tree injection only, as retrieval augmentation is not beneficial under fine-tuning due to training-inference format mismatch. LLM judgment scores reflect o3-adjudicated F1 applied to DRIFT predictions.

Table~\ref{tab:results} presents a comprehensive comparison between DRIFT and state-of-the-art generative ASQP models on Rest15 and Rest16. DRIFT achieves 54.47\% on Rest15 and 64.10\% on Rest16, the best performance among the compared generative ASQP methods, with an average F1 of 59.28\%. DRIFT consistently achieves the highest F1 scores on both benchmarks, outperforming the strongest previous methods. Compared with SimRP~\citep{Jian_Chen_Li_Wang_Zeng_Yao_An_Wu_2025}, DRIFT improves F1 by +1.17 on Rest15 and +0.68 on Rest16. DRIFT also surpasses the recently proposed G2C~\citep{he2026g2c} by +2.07 and +1.30 F1 on Rest15 and Rest16, respectively. Relative to earlier approaches, DRIFT yields even larger gains, including +8.49 over GAS~\citep{zhang-etal-2021-towards-generative}, +5.42 over ILO~\citep{hu-etal-2022-improving-aspect}, and +3.19 over IVLS~\citep{nie2024modeling} on Rest15. These consistent improvements demonstrate that explicitly incorporating dependency structure provides stronger relational reasoning than optimization- and generation-based approaches, including MvP \citep{gou-etal-2023-mvp}, GenDA \citep{wang-etal-2023-generative}, MUL \citep{hu-etal-2023-uncertainty}, OTCL \citep{Li2024OpinionTreeguidedCL}, SimRP, and G2C.

DRIFT maintains a strong precision--recall balance: 54.55\% precision and 54.40\% recall on Rest15, and 62.76\% precision and 65.51\% recall on Rest16. In contrast, IVLS exhibits a clear imbalance (54.46\% precision vs.\ 48.53\% recall on Rest15), reflecting more conservative predictions. LLM-based adjudication further boosts judge-adjusted F1 to 72.41\% on Rest15 and 77.20\% on Rest16, highlighting that many apparent errors under exact-match metrics are semantically correct but differ in surface span boundaries. This validates that dependency-guided reasoning helps the model capture sentiment relations even under boundary variation. To assess robustness beyond a single run, we additionally repeated the PEFT experiments across 5 random seeds, obtaining 54.28 $\pm$ 0.13 F1 on Rest15 and 64.00 $\pm$ 0.08 F1 on Rest16 (full per-seed values in Appendix~\ref{app:seeds}), confirming that the headline result is representative of a low-variance distribution rather than an outlier.

\begin{table*}[t]
\centering
\setlength{\tabcolsep}{18pt}
\renewcommand{\arraystretch}{0.9}
\resizebox{\textwidth}{!}{%
\begin{tabular}{l l lcccccccc}
\toprule
\multirow{2}{*}{\textbf{Methods}} &
\multirow{2}{*}{\textbf{Venue}} &
\multicolumn{3}{c}{\textbf{Rest15}} &
\multicolumn{3}{c}{\textbf{Rest16}} &
\multirow{2}{*}{\textbf{Avg (F1)}} \\
\cmidrule(lr){3-5}\cmidrule(lr){6-8}
 & & Pre & Rec & F1 & Pre & Rec & F1 & \\
\midrule
GAS$^{\dagger}$  & ACL21 & 45.31 & 46.70 & 45.98 & 54.54 & 57.62 & 56.04 & 51.01 \\
Paraphrase       & EMNLP21 & 46.16 & 47.72 & 46.93 & 56.63 & 59.30 & 57.93 & 52.43 \\
DLO                  & EMNLP22 & 47.08 & 49.33 & 48.18 & 57.92 & 61.80 & 59.79 & 53.99 \\
ILO                  & EMNLP22 & 47.08 & 50.38 & 49.05 & 57.58 & 61.17 & 59.32 & 54.19 \\
MvP                 & ACL23 &   -   &   -   & 51.04 &   -   &   -   & 60.39 & 55.72 \\
GenDA             & *SEM23 & 49.74 & 50.29 & 50.01 & 60.08 & 61.70 & 60.88 & 55.45 \\
MUL                  & ACL23 & 49.12 & 50.39 & 49.75 & 59.24 & 61.75 & 60.47 & 55.11 \\
OTCL                 & CSCWD24 & 47.86 & 50.77 & 49.27 & 58.31 & 62.02 & 60.11 & 54.69 \\
IVLS                & Neurocomp24 & \underline{54.46} & 48.53 & 51.28 & 62.69 & 59.75 & 61.04 & 56.16 \\
SimRP              & AAAI25 & 53.12 & \underline{53.50} & \underline{53.30} & \underline{62.74} & 64.12 & \underline{63.42} & \underline{58.36} \\
G2C                & arXiv26 & 51.26 & 53.38 & 52.40 & 61.24 & \underline{64.48} & 62.80 & 57.60 \\

\midrule
\textbf{DRIFT}                    &   & \textbf{54.55} & \textbf{54.40} & \textbf{54.47} & \textbf{62.76} & \textbf{65.51} & \textbf{64.10} & \textbf{59.28} \\

\textit{LLM Judge}        &  & \textit{73.12} & \textit{71.72} & \textit{72.41} & \textit{75.64} & \textit{78.96} & \textit{77.2}  & \textit{74.80} \\
\bottomrule
\end{tabular}%
}
\caption{Performance comparison (\%); $\dagger$ indicates results quoted from GenDA.}
\label{tab:results}
\end{table*}

\subsection{Judge Model Sensitivity Analysis}
\label{sec:judge-models}
To verify robustness to the choice of LLM judge, we repeat the adjudication using GPT-4o and GPT-4o-mini while keeping the prompt and evaluation protocol identical. As shown in Table~\ref{tab:judge_models}, o3 achieves the highest adjusted F1, followed by GPT-4o and GPT-4o-mini. The consistent ranking across datasets confirms that DRIFT’s relative performance and the conclusions drawn are stable, while differences reflect stronger semantic reasoning in more capable LLMs.
\begin{table}[t]
\centering
\footnotesize
\setlength{\tabcolsep}{15pt}
\renewcommand{\arraystretch}{0.8}
\begin{tabular}{lcc}
\toprule
\textbf{Judge Model} & \textbf{Rest15} & \textbf{Rest16} \\
\midrule
o3 (Default) & \textbf{72.41} & \textbf{77.20} \\
GPT-4o       & 63.97 & 71.58 \\
GPT-4o-mini  & 61.94 & 68.43 \\
\bottomrule
\end{tabular}
\caption{F1 (\%) by LLM Judge Model.}
\label{tab:judge_models}
\end{table}

\subsection{Syntactic Format Analysis}

Table~\ref{tab:dep_tree_f1} compares five dependency linearization formats using GPT-4o-mini with one-shot semantic retrieval. Although all formats encode identical syntactic information, performance varies depending on representation style.

More descriptive formats, such as Plain Text and Hierarchical structures, generally achieve stronger performance than compact encodings, suggesting that LLMs benefit from human-readable syntactic representations. However, differences across formats and parsers remain relatively small, indicating that DRIFT is robust to parser and formatting choices. We therefore adopt the spaCy Default representation for subsequent experiments due to its lower token overhead and stronger inference efficiency while maintaining competitive performance.

\begin{table}[t]
\small
\centering
\footnotesize
\setlength{\tabcolsep}{4pt}
\renewcommand{\arraystretch}{0.8}
\begin{tabular}{lcccc}
\toprule
\multirow{2}{*}{\textbf{Format}} & 
\multicolumn{2}{c}{\textbf{spaCy}} & 
\multicolumn{2}{c}{\textbf{Stanza}} \\
\cmidrule(lr){2-3}\cmidrule(lr){4-5}
 & Rest15 & Rest16 & Rest15 & Rest16 \\
\midrule
spaCy (Default) & 34.27 & 43.08 & 35.19 & 43.72 \\
JSON          & 35.11 & 42.94 & \textbf{35.94} & 44.55 \\
CoNLL-U       & 34.55 & \textbf{44.16} & 35.14 & 44.55 \\
Plain Text    & \textbf{35.92} & 43.54 & 35.28 & 44.23 \\
Hierarchical  & 35.85 & 43.94 & 35.51 & \textbf{45.34} \\
\bottomrule
\end{tabular}
\caption{F1 (\%) by dependency parser format.}
\label{tab:dep_tree_f1}
\end{table}

\subsection{ICL Zero/Few-Shot}
\label{sec:fewshot}
\begin{figure}[t]
  \centering
  \begin{minipage}{0.5\columnwidth}
    \centering
    \includegraphics[width=\linewidth]{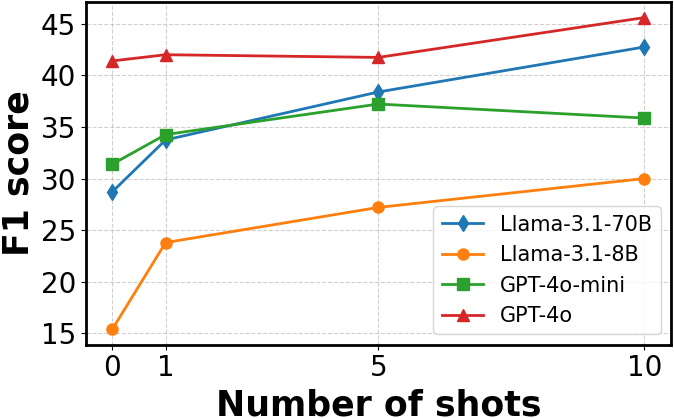}
    \vspace{-8pt}
    \caption*{a: Rest15}
  \end{minipage}\hfill
  \begin{minipage}{0.5\columnwidth}
    \centering
    \includegraphics[width=\linewidth]{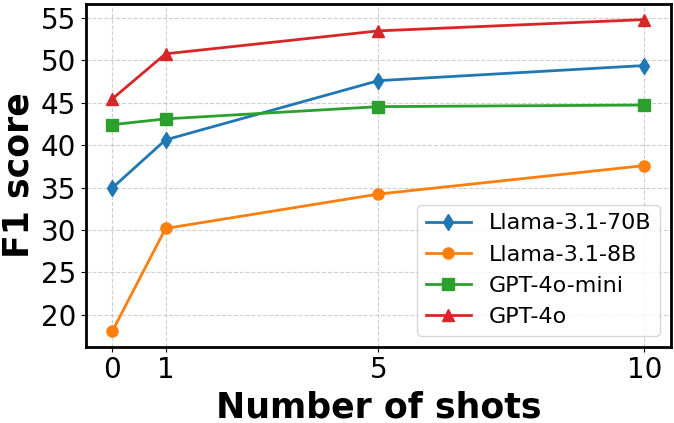}
    \vspace{-8pt}
    \caption*{b: Rest16}
  \end{minipage}
  \vspace{1pt}
  \caption{F1 scores vs.\ number of shots in ICL.}
  \label{fig:shot-scaling}
\end{figure}

Table~\ref{tab:fewshot-results} demonstrates DRIFT's consistent superiority
across zero-shot and few-shot settings. Retrieval selects semantically similar
demonstrations for contextually relevant linguistic patterns, while the
linearized dependency tree explicitly exposes grammatical relations within the
query sentence, enabling the model to identify aspect--opinion links across
long distances and coordination structures that semantic similarity alone
cannot capture. This complementary mechanism explains the smooth, monotonic
performance scaling observed in Figure~\ref{fig:shot-scaling}: as
demonstrations increase, retrieval provides richer task patterns while
dependency structure stabilizes intra-sentence reasoning. Both smaller and
larger models benefit substantially, with DRIFT outperforming the strongest
baseline SimRP across all configurations and backbone
models\footnote{\href{https://openai.com/api/}{\textnormal{OpenAI API:
https://openai.com/api/}}}, confirming that syntax-aware retrieval effectively
leverages additional demonstrations regardless of model scale.

\begin{table*}[t]
\centering
\small
\setlength{\tabcolsep}{14pt}
\renewcommand{\arraystretch}{1}
\resizebox{\textwidth}{!}{%
\begin{tabular}{llcccccccc}
\toprule
\multirow{2}{*}{\textbf{Method}} &
\multirow{2}{*}{\textbf{Backbone}} &
\multicolumn{4}{c}{\textbf{Rest15}} &
\multicolumn{4}{c}{\textbf{Rest16}} \\
\cmidrule(lr){3-6} \cmidrule(lr){7-10}
 & & 0-shot & 1-shot & 5-shot & 10-shot & 0-shot & 1-shot & 5-shot & 10-shot \\
\midrule
\multirow{4}{*}{THOR}
& Llama-3.1-8B   & 7.88  & 8.02  & 8.65  & 10.01 & 9.44  & 10.78 & 11.37 & 11.95 \\
& Llama-3.1-70B  & 17.62 & 22.49 & 26.78 & 31.47 & 27.61 & 30.04 & 34.34 & 35.91 \\
& GPT-4o-mini    & 18.18 & 21.52 & 24.29 & 29.36 & 29.83 & 31.50 & 32.84 & 34.05 \\
& GPT-4o         & 30.79 & 33.37 & 35.12 & 36.81 & 37.23 & 39.22 & 42.05 & 43.57 \\
\midrule
\multirow{4}{*}{LLMs for SA}
& Llama-3.1-8B   & 8.61  & 8.82  & 8.66  & 10.69 & 11.43 & 9.99  & 11.67 & 12.59 \\
& Llama-3.1-70B  & 18.20 & 20.79 & 27.81 & 32.84 & 29.01 & 30.99 & 35.04 & 37.10 \\
& GPT-4o-mini    & 19.80 & 22.16 & 24.93 & 30.11 & 31.43 & 29.42 & 35.84 & 35.45 \\
& GPT-4o         & 34.56 & 35.62 & 36.29 & 37.08 & 39.15 & 39.61 & 43.36 & 45.00 \\
\midrule
\multirow{4}{*}{SimRP}
& Llama-3.1-8B   & 8.88  & 9.26  & 14.13 & 15.26 & 12.55 & 15.46 & 16.71 & 21.60 \\
& Llama-3.1-70B  & 18.12 & 26.83 & 35.56 & 38.19 & 28.71 & 33.73 & 40.91 & 43.81 \\
& GPT-4o-mini    & 19.47 & 26.46 & 33.33 & 35.55 & 31.25 & 39.36 & 40.44 & 43.13 \\
& GPT-4o         & 34.44 & 37.88 & 41.08 & \underline{43.17} & 39.33 & 45.50 & 48.62 & \underline{49.74} \\
\midrule
\multirow{4}{*}{\textbf{DRIFT}}
& Llama-3.1-8B   & 15.41 & 23.80 & 27.21 & 30.01 & 18.09 & 30.18 & 34.23 & 37.58 \\
& Llama-3.1-70B  & 28.73 & 33.75 & 38.40 & 42.76 & 34.96 & 40.61 & 47.58 & 49.36 \\
& GPT-4o-mini    & 31.40 & 34.27 & 37.23 & 35.87 & 42.40 & 43.08 & 44.52 & 44.71 \\
& GPT-4o         & 41.41 & 42.01 & 41.75 & \textbf{45.61} & 45.44 & 50.74 & 53.44 & \textbf{54.77} \\
\bottomrule
\end{tabular}%
}
\caption{F1 (\%) of DRIFT and baseline methods under ICL.}
\label{tab:fewshot-results}
\end{table*}

\subsection{ACOS Dataset Evaluation}

Table~\ref{tab:acos-icl-comparison} reports cross-domain evaluation results on Restaurant-ACOS and Laptop-ACOS under the same ICL setting used for Rest15 and Rest16. DRIFT consistently outperforms standard prompting baselines, including Zero-Shot, Few-Shot, Majority-Vote, and Retrieval-Augmented prompting, except at 0-shot with GPT-4o-mini.

On Restaurant-ACOS, DRIFT achieves the strongest performance, reaching 44.49\% F1 with GPT-4o-mini at 5-shot, and 51.59\% with GPT-4o at 10-shot. Similar improvements are observed on Laptop-ACOS, where DRIFT substantially surpasses retrieval-augmented prompting despite the dataset's larger category space and increased difficulty.

Interestingly, Zero-Shot CoT performs worse than standard Zero-Shot prompting on both datasets, suggesting that generic reasoning alone is insufficient for compositional quadruple extraction. In contrast, dependency-guided prompting provides explicit relational structure that generalizes effectively across domains as demonstrations become available, without task-specific fine-tuning.

\begin{table*}[t]
\centering
\small
\setlength{\tabcolsep}{8pt}
\renewcommand{\arraystretch}{1.1}
\resizebox{\textwidth}{!}{%
\begin{tabular}{llcccccccc}
\toprule
\multirow{2}{*}{\textbf{Method}} &
\multirow{2}{*}{\textbf{Backbone}} &
\multicolumn{4}{c}{\textbf{Rest-ACOS}} &
\multicolumn{4}{c}{\textbf{Laptop-ACOS}} \\
\cmidrule(lr){3-6}\cmidrule(lr){7-10}
 & & 0-shot & 1-shot & 5-shot & 10-shot 
   & 0-shot & 1-shot & 5-shot & 10-shot \\
\midrule
Zero-Shot \citep{NEURIPS2020_1457c0d6}
 & GPT-4o-mini  & \textbf{31.28} & - & - & - & \textbf{11.18} & - & - & - \\
Zero-Shot CoT \citep{NEURIPS2022_8bb0d291}
 & GPT-4o-mini  & 23.01 & - & - & - & 7.56 & - & - & - \\
Few-Shot (N=5) \citep{NEURIPS2020_1457c0d6}
 & GPT-4o-mini  & - & - & 32.76 & - & - & - & 13.69 & - \\
Majority-Vote (N=5, K=8) \citep{li-etal-2025-aligning-black}
 & GPT-4o-mini  & - & - & 34.09 & - & - & - & 15.22 & - \\
Retrieval-Augmented (N=5) \citep{li-etal-2025-aligning-black}
 & GPT-4o-mini  & - & - & 42.15 & - & - & - & 21.87 & - \\
\midrule
\multirow{2}{*}{\textbf{DRIFT}}
 & GPT-4o-mini   
   & \underline{26.69} & \textbf{37.34} & \textbf{44.49} & \textbf{43.59} 
   & \underline{10.19} & \textbf{18.78} & \textbf{23.87} & \textbf{26.00} \\
 & GPT-4o        
   & \textbf{38.73} & \textbf{44.54} & \textbf{47.24} & \textbf{51.59} 
   & \textbf{17.37} & \textbf{21.26} & \textbf{27.75} & \textbf{29.29} \\
\bottomrule
\end{tabular}%
}
\caption{Comparison of DRIFT with different In-Context Learning methods on ACOS benchmarks.}
\label{tab:acos-icl-comparison}
\end{table*}   

\subsection{Ablation Study}
\label{sec:ablation}
Table~\ref{tab:ablation-combined} evaluates the contributions
of dependency trees and retrieval across PEFT and ICL paradigms.
Unless otherwise specified, retrieval-based configurations use one retrieved example ($k=1$) as the few-shot demonstration in both the ICL and PEFT settings, and all ICL experiments are conducted using GPT-4o mini as the underlying LLM.
In PEFT, dependency trees alone (w/o retrieval) yield the
strongest result at 54.47\% and 64.10\%, while removing both
components drops performance by $-$1.32 on Rest15 and $-$1.30
on Rest16 relative to full DRIFT (syntax + retrieval: 51.01\%
and 63.17\%). The drop when adding retrieval to PEFT
(54.47\%~$\rightarrow$~51.01\% on Rest15) is expected by
design: the model is trained exclusively on syntax-augmented
prompts without retrieved demonstrations, so injecting
retrieved examples at inference time creates a distribution
mismatch that introduces noise. This confirms that
dependency-guided reasoning is sufficient and self-contained
under fine-tuning, while retrieval augmentation is better
suited to ICL where no parameter updates occur.

In ICL, both components become substantially more critical.
Removing both from the ICL setting causes drops of $-$9.38
on Rest15 and $-$13.38 on Rest16 relative to full DRIFT ICL
(34.27\% and 43.08\%), far exceeding the corresponding PEFT
drops of $-$1.32 and $-$1.30. Examining each component
individually, removing dependency trees while keeping
retrieval yields 34.79\% and 41.52\%, whereas removing
retrieval while keeping dependency trees yields 31.40\% and
42.40\%, indicating that retrieval contributes more in
few-shot settings on Rest15 (+3.39 marginal gain from
31.40\% to 34.79\%), while dependency trees provide a larger
absolute improvement under PEFT (+4.34 from 50.13\% to
54.47\%). These results reveal complementary roles: retrieval
supports inter-instance generalization through semantically
relevant demonstrations, while dependency trees guide
intra-sentence aspect--opinion reasoning. DRIFT's
effectiveness arises from the interaction of both components
across learning paradigms.

\begin{table}[t]
\centering
\small
\setlength{\tabcolsep}{6pt}
\renewcommand{\arraystretch}{0.8}
\resizebox{\columnwidth}{!}{%
\begin{tabular}{c c cc cc}
\toprule
\multirow{2}{*}{\textbf{Dep. Tree}} &
\multirow{2}{*}{\textbf{Retrieval}} &
\multicolumn{2}{c}{\textbf{PEFT}} &
\multicolumn{2}{c}{\textbf{ICL}} \\
\cmidrule(lr){3-4}\cmidrule(lr){5-6}
 & & Rest15 & Rest16 & Rest15 & Rest16 \\
\midrule
\xmark & \xmark & 49.69 & 61.87 & 24.89 & 29.70 \\
\xmark & \cmark & 50.13 & 61.34 & \textbf{34.79} & 41.52 \\
\cmark & \xmark & \textbf{54.47} & \textbf{64.10} & 31.40 & 42.40 \\
\cmark & \cmark & 51.01 & 63.17 & 34.27 & \textbf{43.08} \\
\bottomrule
\end{tabular}%
}
\caption{Ablations in PEFT and ICL.}
\label{tab:ablation-combined}
\end{table}

\newcommand{\placeholder}[0]{\textbf{?}}

\subsection{Dependency Distance Analysis}
\label{sec:dep-complexity}

To test whether dependency injection's benefit is concentrated in
structurally difficult cases, we partition test instances by the shortest
dependency-tree distance between the gold aspect and opinion spans (spaCy
shortest path): d=1--2, d=3--4, and d$\geq$5, and
compare zero-shot exact-match F1 with and without dependency injection
(GPT-4o-mini, no retrieval) within each bucket on Rest15 and Rest16.

Table~\ref{tab:dep-distance} shows dependency injection improves F1 in
five of six dataset-bucket combinations, with the largest gains at short
and medium distances on Rest16 (+8.66, +5.52) and smaller, consistent gains
across Rest15. At the longest range (d$\geq$5), the effect is small and
inconsistent (+0.72 on Rest15, $-$4.11 on Rest16), where this bucket also
has the fewest instances. We interpret this as a broad, concentrated positive effect rather than a monotonic benefit with distance.

\begin{table}[t]
\centering
\small
\setlength{\tabcolsep}{6pt}
\renewcommand{\arraystretch}{0.6}
\begin{tabular}{lcccc}
\toprule
\textbf{Dataset} & \textbf{Bucket} & \textbf{Vanilla F1} & \textbf{+Dep F1} & \textbf{$\Delta$} \\
\midrule
\multirow{3}{*}{Rest15}
 & 1--2    & 37.57 & 39.69 & +2.12 \\
 & 3--4    & 27.04 & 27.62 & +0.58 \\
 & $\geq$5 & 16.67 & 17.39 & +0.72 \\
\midrule
\multirow{3}{*}{Rest16}
 & 1--2    & 39.42 & 48.08 & +8.66 \\
 & 3--4    & 31.87 & 37.39 & +5.52 \\
 & $\geq$5 & 21.85 & 17.74 & $-$4.11 \\
\bottomrule
\end{tabular}
\caption{Dependency-distance analysis.}
\label{tab:dep-distance}
\end{table}

\subsection{Span Boundary Variation Analysis}
\label{sec:span-boundary}

Model predictions often preserve the same aspect--sentiment semantics as the reference while differing in span boundaries or attachment granularity (e.g., \textit{delivers} attaching to \textit{spot} vs.\ \textit{modern Japanese}; \textit{GREAT} vs.\ \textit{GREAT TIME}), reflecting annotation flexibility rather than genuine errors.

To assess adjudication reliability, we randomly sampled 100 cases where exact-match marked a prediction incorrect but the o3 adjudicator judged it semantically equivalent. We evaluated these blindly and reached 84\% agreement with the adjudicator, indicating most disagreements arise from annotation variation rather than incorrect predictions. Representative examples are in Table~\ref{tab:qualitative-examples}, with detailed analysis in Appendix~\ref{app:qualitative}.

Challenging cases remain for sentences with multiple coordinated aspect--opinion pairs, nested clauses, and highly implicit sentiment requiring pragmatic or domain-specific knowledge. Dependency-guided prompting substantially improves relational reasoning, though complex discourse-level semantics remain an open challenge.

\subsection{Inference Efficiency of DRIFT}
To evaluate the practical efficiency of DRIFT, we measure inference latency,
token usage, and computational complexity on 100 randomly selected Rest15
examples using GPT-4o mini as the underlying LLM with one retrieved example
($k=1$) as the few-shot demonstration. Table~\ref{tab:inference-cost} shows that the Prompt + Prediction step dominates inference, averaging 1.306 seconds per instance and approximately 590 tokens, while dependency parsing and few-shot retrieval add less than 0.02 seconds of overhead, keeping the total prompt length below 800 tokens per example.

From a computational perspective, dependency parsing and prompt-based prediction 
operate with effectively constant-time complexity under fixed prompt lengths, 
whereas retrieval scales linearly with the number of indexed examples due to 
embedding similarity search. Overall, the results indicate that DRIFT improves 
extraction performance with modest computational and token overhead, supporting 
its practical applicability for real-world ASQP settings.
\begin{table}[t]
\centering
\footnotesize
\setlength{\tabcolsep}{5pt}
\renewcommand{\arraystretch}{1.15}
\resizebox{\columnwidth}{!}{%
\begin{tabular}{lcccc}
\toprule
\textbf{Component} &
\textbf{Time (s)} &
\textbf{Std} &
\textbf{Avg. Tokens} &
\textbf{Complexity} \\
\midrule
  Dependency Parsing & 0.015 & 0.004 & 109.0 & $\mathcal{O}(1)$ \\
  Few-shot Retrieval & 0.014 & 0.004 & 81.7 & $\mathcal{O}(n)$ \\
  Prompt + Prediction & 1.306 & 0.923 & 590.1 & $\mathcal{O}(1)$ \\
\midrule
  \textbf{Total} & \textbf{1.335} & \textbf{0.923} & \textbf{780.8} & \textbf{$\mathcal{O}(n)$} \\
\bottomrule
\end{tabular}%
}
\caption{Inference cost analysis of DRIFT.}
\label{tab:inference-cost}
\end{table}

\section{Conclusion}

We presented DRIFT, a structure-grounded framework for ASQP that integrates dependency-guided prompting, retrieval-augmented inference, and syntax-aware LLM adjudication. DRIFT achieves the best performance among compared generative ASQP methods on the Rest15 and Rest16 benchmarks under both PEFT and ICL settings, and on the Restaurant-ACOS and Laptop-ACOS benchmarks under ICL. Our analyses show that dependency-guided prompting and retrieval augmentation provide complementary benefits, while syntax-aware semantic adjudication offers more reliable evaluation by reducing sensitivity to span-boundary variation. These findings demonstrate that explicitly incorporating dependency structure as reasoning context improves both extraction quality and evaluation reliability without requiring architectural modifications. Overall, our results highlight the value of explicit dependency-guided reasoning for LLM-based structured sentiment analysis and suggest a promising direction for future structure-aware information extraction and structured prediction.

\section*{Limitations}

DRIFT relies on external dependency parsers and is therefore sensitive to parsing errors, particularly in noisy or informal text. Retrieval quality may also degrade performance under domain shift. In addition, benchmark datasets contain annotation inconsistencies and span ambiguities that are beyond our control and partially motivate our semantic adjudication framework. Although semantic adjudication improves evaluation robustness, it incurs additional computational cost and depends on the reliability of the underlying LLM. 
Finally, our experiments focus primarily on GPT-based models; future work will investigate multilingual settings and a broader range of open-source LLMs.

\section*{Ethical Considerations}

This work uses publicly available and anonymized benchmark datasets (Rest15, Rest16, and ACOS) that do not contain personally identifiable information. Large language models are employed for ASQP prediction and semantic adjudication, which may introduce model-specific biases and inconsistencies. To mitigate this issue, semantic adjudication is used only as a complementary analysis alongside standard exact-match evaluation.

Dependency parsing and retrieval components may additionally inherit biases from pretrained models and their training corpora. Consequently, caution should be exercised when applying DRIFT to sensitive or high-stakes domains without further validation. All experiments follow the documented terms of use of the employed models and resources, and implementation details are provided to support reproducibility and further research.


\bibliographystyle{acl_natbib}
\bibliography{references}

@inproceedings{Zhang2021ASQP,
  title = {Aspect Sentiment Quad Prediction as Paraphrase Generation},
  author = {Wenxuan Zhang and Yang Deng and Xin Li and Yifei Yuan and Lidong Bing and Wai Lam},
  booktitle = {Proceedings of the 2021 Conference on Empirical Methods in Natural Language Processing (EMNLP)},
  year = {2021},
  pages = {9209--9219},
  url = {https://aclanthology.org/2021.emnlp-main.726/},
  doi = {10.18653/v1/2021.emnlp-main.726},
  publisher = {Association for Computational Linguistics}
}

@inproceedings{Wan2020TASD,
  title = {Target-Aspect-Sentiment Joint Detection for Aspect-Based Sentiment Analysis},
  author = {Hai Wan and Yufei Yang and Jianfeng Du and Yanan Liu and Kunxun Qi and Jeff Z. Pan},
  booktitle = {Proceedings of the 34th {AAAI} Conference on Artificial Intelligence (AAAI-20)},
  year = {2020},
  pages = {9122--9129},
  doi = {10.1609/aaai.v34i05.6447},
  publisher = {Association for the Advancement of Artificial Intelligence}
}

@inproceedings{Mao2021JointASQP,
  title = {A Joint Training Dual-{MRC} Framework for Aspect Based Sentiment Analysis},
  author = {Yue Mao and Yi Shen and Chao Yu and Longjun Cai},
  booktitle = {Proceedings of the 35th {AAAI} Conference on Artificial Intelligence (AAAI-21)},
  year = {2021},
  pages = {13543--13550},
  publisher = {AAAI Press},
  doi = {10.1609/aaai.v35i15.17597}
}

@inproceedings{Cai2021ACOS,
  title = {Aspect-Category-Opinion-Sentiment Quadruple Extraction with Implicit Aspects and Opinions},
  author = {Hongjie Cai and Rui Xia and Jianfei Yu},
  booktitle = {Proceedings of the 59th Annual Meeting of the Association for Computational Linguistics (ACL 2021)},
  year = {2021},
  pages = {340--350},
  address = {Online},
  publisher = {Association for Computational Linguistics},
  doi= {10.18653/v1/2021.acl-long.29}
}

@article{Zhang2024SDGCN,
  author = {Fan Zhang and Wenbin Zheng and Yujie Yang},
  title = {Graph Convolutional Network with Syntactic Dependency for Aspect-Based Sentiment Analysis},
  journal = {International Journal of Computational Intelligence Systems},
  volume = {17},
  number = {1},
  pages = {37},
  year = {2024},
  doi = {10.1007/s44196-024-00419-6}
}

@article{Liang2022BSGCN,
  author = {Bin Liang and Hang Su and Lin Gui and Erik Cambria and Ruifeng Xu},
  title = {Aspect-Based Sentiment Analysis via Affective Knowledge Enhanced Graph Convolutional Networks},
  journal = {Knowledge-Based Systems},
  volume = {235},
  pages = {107643},
  year = {2022},
  doi = {10.1016/j.knosys.2021.107643}
}

@inproceedings{NEURIPS2020_1457c0d6,
 author = {Brown, Tom and Mann, Benjamin and Ryder, Nick and Subbiah, Melanie and Kaplan, Jared D and Dhariwal, Prafulla and Neelakantan, Arvind and Shyam, Pranav and Sastry, Girish and Askell, Amanda and Agarwal, Sandhini and Herbert-Voss, Ariel and Krueger, Gretchen and Henighan, Tom and Child, Rewon and Ramesh, Aditya and Ziegler, Daniel and Wu, Jeffrey and Winter, Clemens and Hesse, Chris and Chen, Mark and Sigler, Eric and Litwin, Mateusz and Gray, Scott and Chess, Benjamin and Clark, Jack and Berner, Christopher and McCandlish, Sam and Radford, Alec and Sutskever, Ilya and Amodei, Dario},
 booktitle = {Advances in Neural Information Processing Systems},
 editor = {H. Larochelle and M. Ranzato and R. Hadsell and M.F. Balcan and H. Lin},
 pages = {1877--1901},
 publisher = {Curran Associates, Inc.},
 title = {Language Models are Few-Shot Learners},
 url = {https://proceedings.neurips.cc/paper_files/paper/2020/file/1457c0d6bfcb4967418bfb8ac142f64a-Paper.pdf},
 volume = {33},
 year = {2020}
}

@inproceedings{scaria-etal-2024-instructabsa,
    title = "{I}nstruct{ABSA}: Instruction Learning for Aspect Based Sentiment Analysis",
    author = "Scaria, Kevin  and
      Gupta, Himanshu  and
      Goyal, Siddharth  and
      Sawant, Saurabh  and
      Mishra, Swaroop  and
      Baral, Chitta",
    editor = "Duh, Kevin  and
      Gomez, Helena  and
      Bethard, Steven",
    booktitle = "Proceedings of the 2024 Conference of the North American Chapter of the Association for Computational Linguistics: Human Language Technologies (Volume 2: Short Papers)",
    month = jun,
    year = "2024",
    address = "Mexico City, Mexico",
    publisher = "Association for Computational Linguistics",
    url = "https://aclanthology.org/2024.naacl-short.63/",
    doi = "10.18653/v1/2024.naacl-short.63",
    pages = "720--736"
}

@inproceedings{zheng-etal-2024-instruction,
    title = "Instruction Tuning with Retrieval-based Examples Ranking for Aspect-based Sentiment Analysis",
    author = "Zheng, Guangmin  and
      Wang, Jin  and
      Yu, Liang-Chih  and
      Zhang, Xuejie",
    editor = "Ku, Lun-Wei  and
      Martins, Andre  and
      Srikumar, Vivek",
    booktitle = "Findings of the Association for Computational Linguistics: ACL 2024",
    month = aug,
    year = "2024",
    address = "Bangkok, Thailand",
    publisher = "Association for Computational Linguistics",
    url = "https://aclanthology.org/2024.findings-acl.284/",
    doi = "10.18653/v1/2024.findings-acl.284",
    pages = "4777--4788"
}

@inproceedings {Jian_Chen_Li_Wang_Zeng_Yao_An_Wu_2025, 
title={SimRP: Syntactic and Semantic Similarity Retrieval Prompting Enhances Aspect Sentiment Quad Prediction},
volume={39}, 
url={https://ojs.aaai.org/index.php/AAAI/article/view/34601},
doi = {10.1609/aaai.v39i23.34601},
author={Jian, Zhongquan and Chen, Yanhao and Li, Jiajian and Wang, Shaopan and Zeng, Xiangjian and Yao, Junfeng and An, Xinying and Wu, Qingqiang}, 
year={2025}, 
booktitle = {Proceedings of the 39th AAAI Conference on Artificial Intelligence (AAAI 2025)},
month={Apr.}, 
pages={24248-24256} 
}

@inproceedings{Zheng2023JudgingLLM,
  author = {Lianmin Zheng and Wei-Lin Chiang and Ying Sheng and Siyuan Zhuang and Zhanghao Wu and Yonghao Zhuang and Zi Lin and Zhuohan Li and Dacheng Li and Eric P. Xing and Hao Zhang and Joseph E. Gonzalez and Ion Stoica},
  title = {Judging LLM-as-a-Judge with MT-Bench and Chatbot Arena},
  booktitle = {Neural Information Processing Systems (NeurIPS) 2023 – Datasets and Benchmarks Track},
  year = {2023},
  doi= {10.48550/arXiv.2306.05685}
}

@inproceedings{Liu2023GEVAL,
  author = {Yang Liu and Dan Iter and Yichong Xu and Shuohang Wang and Ruochen Xu and Chenguang Zhu},
  title = {G-Eval: NLG Evaluation using GPT-4 with Better Human Alignment},
  booktitle = {Proceedings of the 2023 Conference on Empirical Methods in Natural Language Processing (EMNLP)},
  year = {2023},
  pages = {2511--2522},
  publisher = {Association for Computational Linguistics},
  address = {Singapore},
  doi = {10.18653/v1/2023.emnlp-main.153}
}

@article{Hua2024FTSOBP,
  author = {Yan Cathy Hua and Paul Denny and J\"org Wicker and Katerina Ta\u{s}kov{a}},
  title = {Data-Efficient Adaptation and a Novel Evaluation Method for Aspect-based Sentiment Analysis},
  journal = {arXiv preprint arXiv:2511.03034},
  year = {2025},
  doi = {10.48550/arXiv.2511.03034}
}

@inproceedings{yang-etal-2025-single,
    title = "Single Ground Truth Is Not Enough: Adding Flexibility to Aspect-Based Sentiment Analysis Evaluation",
    author = "Yang, Soyoung  and
      Cho, Hojun  and
      Lee, Jiyoung  and
      Yoon, Sohee  and
      Choi, Edward  and
      Choo, Jaegul  and
      Cho, Won Ik",
    editor = "Chiruzzo, Luis  and
      Ritter, Alan  and
      Wang, Lu",
    booktitle = "Proceedings of the 2025 Conference of the Nations of the Americas Chapter of the Association for Computational Linguistics: Human Language Technologies (Volume 1: Long Papers)",
    month = apr,
    year = "2025",
    address = "Albuquerque, New Mexico",
    publisher = "Association for Computational Linguistics",
    url = "https://aclanthology.org/2025.naacl-long.603/",
    doi = "10.18653/v1/2025.naacl-long.603",
    pages = "12071--12096",
    ISBN = "979-8-89176-189-6"
}

@inproceedings{zhang-etal-2023-span,
    title = "Span-level Aspect-based Sentiment Analysis via Table Filling",
    author = "Zhang, Mao  and
      Zhu, Yongxin  and
      Liu, Zhen  and
      Bao, Zhimin  and
      Wu, Yunfei  and
      Sun, Xing  and
      Xu, Linli",
    editor = "Rogers, Anna  and
      Boyd-Graber, Jordan  and
      Okazaki, Naoaki",
    booktitle = "Proceedings of the 61st Annual Meeting of the Association for Computational Linguistics (Volume 1: Long Papers)",
    month = jul,
    year = "2023",
    address = "Toronto, Canada",
    publisher = "Association for Computational Linguistics",
    url = "https://aclanthology.org/2023.acl-long.515/",
    doi = "10.18653/v1/2023.acl-long.515",
    pages = "9273--9284"
}

@inproceedings{Zhang2019GCN,
  author = {Chen Zhang and Qiuchi Li and Dawei Song},
  title = {Aspect-based Sentiment Classification with Aspect-specific Graph Convolutional Networks},
  booktitle = {Proceedings of the 2019 Conference on Empirical Methods in Natural Language Processing (EMNLP)},
  year = {2019},
  pages = {4568--4578},
  publisher = {Association for Computational Linguistics},
  doi = {10.18653/v1/D19-1464}
}

@inproceedings{HuangCarley2020Attention,
  author = {Binxuan Huang and Kathleen M. Carley},
  title = {Syntax-Aware Aspect Level Sentiment Classification with Graph Attention Networks},
  booktitle = {Proceedings of the 2019 Conference on Empirical Methods in Natural Language Processing (EMNLP)},
  year = {2019},
  pages = {5469--5477},
  publisher = {Association for Computational Linguistics},
  doi = {10.18653/v1/D19-1549}
}

@inproceedings{Devlin2019,
  author = {Jacob Devlin and Ming-Wei Chang and Kenton Lee and Kristina Toutanova},
  title = {{BERT}: Pre-training of Deep Bidirectional Transformers for Language Understanding},
  booktitle = {Proceedings of the 2019 Conference of the North American Chapter of the Association for Computational Linguistics (NAACL)},
  year = {2019},
  pages = {4171--4186},
  publisher = {Association for Computational Linguistics},
  doi = {10.18653/v1/N19-1423}
}

@inproceedings{Sun2019,
  author = {Chi Sun and Luyao Huang and Xipeng Qiu},
  title = {Utilizing {BERT} for Aspect-Based Sentiment Analysis via Constructing Auxiliary Sentence},
  booktitle = {Proceedings of the 2019 Conference of the North American Chapter of the Association for Computational Linguistics (NAACL)},
  year = {2019},
  pages = {380--385},
  publisher = {Association for Computational Linguistics},
  doi = {10.18653/v1/N19-1035}
}

@inproceedings{scialom-etal-2022-fine,
    title = "Fine-tuned Language Models are Continual Learners",
    author = "Scialom, Thomas  and
      Chakrabarty, Tuhin  and
      Muresan, Smaranda",
    editor = "Goldberg, Yoav  and
      Kozareva, Zornitsa  and
      Zhang, Yue",
    booktitle = "Proceedings of the 2022 Conference on Empirical Methods in Natural Language Processing",
    month = dec,
    year = "2022",
    address = "Abu Dhabi, United Arab Emirates",
    publisher = "Association for Computational Linguistics",
    url = "https://aclanthology.org/2022.emnlp-main.410/",
    doi = "10.18653/v1/2022.emnlp-main.410",
    pages = "6107--6122"
}

@inproceedings{guu2020realm,
  title        = {REALM: Retrieval-Augmented Language Model Pre-Training},
  author       = {Guu, Kelvin and Lee, Kenton and Tung, Zora and Pasupat, Panupong and Chang, Ming-Wei},
  booktitle    = {Proceedings of the 37th International Conference on Machine Learning (ICML)},
  pages        = {3929--3938},
  year         = {2020},
  publisher    = {PMLR},
  series       = {Proceedings of Machine Learning Research},
  volume       = {119},
  url          = {https://proceedings.mlr.press/v119/guu20a.html}
}

@inproceedings{wei2022chainofThought,
  title     = {Chain-of-Thought Prompting Elicits Reasoning in Large Language Models},
  author    = {Wei, Jason and Wang, Xuezhi and Schuurmans, Dale and Bosma, Maarten and Ichter, Brian and Xia, Fei and Chi, Ed H. and Le, Quoc V. and Zhou, Denny},
  booktitle = {Advances in Neural Information Processing Systems 35 (NeurIPS 2022)},
  pages     = {24824--24837},
  year      = {2022},
  url = {https://doi.org/10.48550/arXiv.2201.11903},
  publisher = {Curran Associates, Inc.}
}

@inproceedings{zhang-etal-2021-towards-generative,
    title = "Towards Generative Aspect-Based Sentiment Analysis",
    author = "Zhang, Wenxuan  and
      Li, Xin  and
      Deng, Yang  and
      Bing, Lidong  and
      Lam, Wai",
    editor = "Zong, Chengqing  and
      Xia, Fei  and
      Li, Wenjie  and
      Navigli, Roberto",
    booktitle = "Proceedings of the 59th Annual Meeting of the Association for Computational Linguistics and the 11th International Joint Conference on Natural Language Processing (Volume 2: Short Papers)",
    month = aug,
    year = "2021",
    address = "Online",
    publisher = "Association for Computational Linguistics",
    url = "https://aclanthology.org/2021.acl-short.64/",
    doi = "10.18653/v1/2021.acl-short.64",
    pages = "504--510"
}

@inproceedings{hu-etal-2022-improving-aspect,
    title = "Improving Aspect Sentiment Quad Prediction via Template-Order Data Augmentation",
    author = "Hu, Mengting  and
      Wu, Yike  and
      Gao, Hang  and
      Bai, Yinhao  and
      Zhao, Shiwan",
    editor = "Goldberg, Yoav  and
      Kozareva, Zornitsa  and
      Zhang, Yue",
    booktitle = "Proceedings of the 2022 Conference on Empirical Methods in Natural Language Processing",
    month = dec,
    year = "2022",
    address = "Abu Dhabi, United Arab Emirates",
    publisher = "Association for Computational Linguistics",
    url = "https://aclanthology.org/2022.emnlp-main.538/",
    doi = "10.18653/v1/2022.emnlp-main.538",
    pages = "7889--7900"
}

@inproceedings{gou-etal-2023-mvp,
    title = "{M}v{P}: Multi-view Prompting Improves Aspect Sentiment Tuple Prediction",
    author = "Gou, Zhibin  and
      Guo, Qingyan  and
      Yang, Yujiu",
    editor = "Rogers, Anna  and
      Boyd-Graber, Jordan  and
      Okazaki, Naoaki",
    booktitle = "Proceedings of the 61st Annual Meeting of the Association for Computational Linguistics (Volume 1: Long Papers)",
    month = jul,
    year = "2023",
    address = "Toronto, Canada",
    publisher = "Association for Computational Linguistics",
    url = "https://aclanthology.org/2023.acl-long.240/",
    doi = "10.18653/v1/2023.acl-long.240",
    pages = "4380--4397"
}

@inproceedings{wang-etal-2023-generative,
    title = "Generative Data Augmentation for Aspect Sentiment Quad Prediction",
    author = "Wang, An  and
      Jiang, Junfeng  and
      Ma, Youmi  and
      Liu, Ao  and
      Okazaki, Naoaki",
    editor = "Palmer, Alexis  and
      Camacho-collados, Jose",
    booktitle = "Proceedings of the 12th Joint Conference on Lexical and Computational Semantics (*SEM 2023)",
    month = jul,
    year = "2023",
    address = "Toronto, Canada",
    publisher = "Association for Computational Linguistics",
    url = "https://aclanthology.org/2023.starsem-1.12/",
    doi = "10.18653/v1/2023.starsem-1.12",
    pages = "128--140"
}

@inproceedings{hu-etal-2023-uncertainty,
    title = "Uncertainty-Aware Unlikelihood Learning Improves Generative Aspect Sentiment Quad Prediction",
    author = "Hu, Mengting  and
      Bai, Yinhao  and
      Wu, Yike  and
      Zhang, Zhen  and
      Zhang, Liqi  and
      Gao, Hang  and
      Zhao, Shiwan  and
      Huang, Minlie",
    editor = "Rogers, Anna  and
      Boyd-Graber, Jordan  and
      Okazaki, Naoaki",
    booktitle = "Findings of the Association for Computational Linguistics: ACL 2023",
    month = jul,
    year = "2023",
    address = "Toronto, Canada",
    publisher = "Association for Computational Linguistics",
    url = "https://aclanthology.org/2023.findings-acl.851/",
    doi = "10.18653/v1/2023.findings-acl.851",
    pages = "13481--13494"
}

@article{Li2024OpinionTreeguidedCL,
  title={Opinion-Tree-guided Contrastive Learning for Aspect Sentiment Quadruple Prediction},
  author={Zhijun Li and Zhenyu Yang and Yiwen Li and Xiaoyang Li},
  journal={2024 27th International Conference on Computer Supported Cooperative Work in Design (CSCWD)},
  year={2024},
  pages={1944-1951},
  url={https://api.semanticscholar.org/CorpusID:271090765}
}

@inproceedings{fei-etal-2023-reasoning,
    title = "Reasoning Implicit Sentiment with Chain-of-Thought Prompting",
    author = "Fei, Hao  and
      Li, Bobo  and
      Liu, Qian  and
      Bing, Lidong  and
      Li, Fei  and
      Chua, Tat-Seng",
    editor = "Rogers, Anna  and
      Boyd-Graber, Jordan  and
      Okazaki, Naoaki",
    booktitle = "Proceedings of the 61st Annual Meeting of the Association for Computational Linguistics (Volume 2: Short Papers)",
    month = jul,
    year = "2023",
    address = "Toronto, Canada",
    publisher = "Association for Computational Linguistics",
    url = "https://aclanthology.org/2023.acl-short.101/",
    doi = "10.18653/v1/2023.acl-short.101",
    pages = "1171--1182"
}

@inproceedings{zhang-etal-2024-sentiment,
    title = "Sentiment Analysis in the Era of Large Language Models: A Reality Check",
    author = "Zhang, Wenxuan  and
      Deng, Yue  and
      Liu, Bing  and
      Pan, Sinno  and
      Bing, Lidong",
    editor = "Duh, Kevin  and
      Gomez, Helena  and
      Bethard, Steven",
    booktitle = "Findings of the Association for Computational Linguistics: NAACL 2024",
    month = jun,
    year = "2024",
    address = "Mexico City, Mexico",
    publisher = "Association for Computational Linguistics",
    url = "https://aclanthology.org/2024.findings-naacl.246/",
    doi = "10.18653/v1/2024.findings-naacl.246",
    pages = "3881--3906"
}

@inproceedings{Pontiki2015,
  author = {Maria Pontiki and Dimitris Galanis and Haris Papageorgiou and Suresh Manandhar and Ion Androutsopoulos},
  title = {{SemEval-2015 Task 12}: Aspect Based Sentiment Analysis},
  booktitle = {Proceedings of the 9th International Workshop on Semantic Evaluation (SemEval 2015)},
  year = {2015},
  pages = {486--495},
  publisher = {Association for Computational Linguistics},
  address = {Denver, Colorado},
  doi = {10.18653/v1/S15-2082}
}

@inproceedings{Pontiki2016,
  author = {Maria Pontiki and Dimitris Galanis and Haris Papageorgiou and Ion Androutsopoulos and Suresh Manandhar and Mohammad AL-Smadi and Mahmoud Al-Ayyoub and Yanyan Zhao and Bing Qin and Orph{\'e}e De Clercq and V{\'e}ronique Hoste and Marianna Apidianaki and Xavier Tannier and Natalia Loukachevitch and Evgeniy Kotelnikov and Nuria Bel and Salud Mar{\'i}a Jim{\'e}nez-Zafra and G{\"u}l{\c{s}}en Ery{\u{g}}it},
  title = {{SemEval-2016 Task 5}: Aspect Based Sentiment Analysis},
  booktitle = {Proceedings of the 10th International Workshop on Semantic Evaluation (SemEval-2016)},
  year = {2016},
  pages = {19--30},
  publisher = {Association for Computational Linguistics},
  address = {San Diego, California},
  doi = {10.18653/v1/S16-1002}
}

@article{Nie2024Modeling,
    title = {Modeling implicit variable and latent structure for aspect-based sentiment quadruple extraction},
    author = {Yu Nie and Jianming Fu and Yu Zhang and Chao Li},
    journal = {Neurocomputing},
    volume = {586},
    pages = {127642},
    year = {2024},
    issn = {0925-2312},
    doi = {10.1016/j.neucom.2024.127642}
}

@article{lora2022hu,
  title={Lora: Low-rank adaptation of large language models.},
  author={Hu, Edward J and Shen, Yelong and Wallis, Phillip and Allen-Zhu, Zeyuan and Li, Yuanzhi and Wang, Shean and Wang, Lu and Chen, Weizhu and others},
  journal={ICLR},
  volume={1},
  number={2},
  pages={3},
  year={2022},
  url = "https://openreview.net/forum?id=nZeVKeeFYf9"
}

@inproceedings{zhang-etal-2024-self-training,
    title = "Self-Training with Pseudo-Label Scorer for Aspect Sentiment Quad Prediction",
    author = "Zhang, Yice  and
      Zeng, Jie  and
      Hu, Weiming  and
      Wang, Ziyi  and
      Chen, Shiwei  and
      Xu, Ruifeng",
    editor = "Ku, Lun-Wei  and
      Martins, Andre  and
      Srikumar, Vivek",
    booktitle = "Proceedings of the 62nd Annual Meeting of the Association for Computational Linguistics (Volume 1: Long Papers)",
    month = aug,
    year = "2024",
    address = "Bangkok, Thailand",
    publisher = "Association for Computational Linguistics",
    url = "https://aclanthology.org/2024.acl-long.640/",
    doi = "10.18653/v1/2024.acl-long.640",
    pages = "11862--11875"
}

@inproceedings{li-etal-2025-aligning-black,
    title = "Aligning Black-Box {LLM}s for Aspect Sentiment Quad Prediction",
    author = "Li, Shichen  and
      Zhang, Jiawei  and
      Wang, Zhongqing  and
      Li, Peifeng",
    editor = "Christodoulopoulos, Christos  and
      Chakraborty, Tanmoy  and
      Rose, Carolyn  and
      Peng, Violet",
    booktitle = "Findings of the Association for Computational Linguistics: EMNLP 2025",
    month = nov,
    year = "2025",
    address = "Suzhou, China",
    publisher = "Association for Computational Linguistics",
    url = "https://aclanthology.org/2025.findings-emnlp.53/",
    doi = "10.18653/v1/2025.findings-emnlp.53",
    pages = "1012--1025",
    ISBN = "979-8-89176-335-7"
}

@inproceedings{su-etal-2025-unified,
    title = "Unified Grid Tagging Scheme for Aspect Sentiment Quad Prediction",
    author = "Su, Guixin  and
      Zhang, Yongcheng  and
      Wang, Tongguan  and
      Wu, Mingmin  and
      Sha, Ying",
    editor = "Rambow, Owen  and
      Wanner, Leo  and
      Apidianaki, Marianna  and
      Al-Khalifa, Hend  and
      Eugenio, Barbara Di  and
      Schockaert, Steven",
    booktitle = "Proceedings of the 31st International Conference on Computational Linguistics",
    month = jan,
    year = "2025",
    address = "Abu Dhabi, UAE",
    publisher = "Association for Computational Linguistics",
    url = "https://aclanthology.org/2025.coling-main.269/",
    pages = "3997--4010"
}

@inproceedings{chen-etal-2025-using,
    title = "Using Review Combination and Pseudo-Tokens for Aspect Sentiment Quad Prediction",
    author = "Chen, Jiazhou  and
      Jia, Xu  and
      Guo, RuiQiang",
    editor = "Chiruzzo, Luis  and
      Ritter, Alan  and
      Wang, Lu",
    booktitle = "Findings of the Association for Computational Linguistics: NAACL 2025",
    month = apr,
    year = "2025",
    address = "Albuquerque, New Mexico",
    publisher = "Association for Computational Linguistics",
    url = "https://aclanthology.org/2025.findings-naacl.214/",
    doi = "10.18653/v1/2025.findings-naacl.214",
    pages = "3872--3883",
    ISBN = "979-8-89176-195-7"
}

@inproceedings{chen-etal-2025-s2it,
    title = "$\mathcal{S}^2${IT}: Stepwise Syntax Integration Tuning for Large Language Models in Aspect Sentiment Quad Prediction",
    author = "Chen, Bingfeng  and
      Qiu, Chenjie  and
      Xie, Yifeng  and
      Xu, Boyan  and
      Cai, Ruichu  and
      Hao, Zhifeng",
    editor = "Chiruzzo, Luis  and
      Ritter, Alan  and
      Wang, Lu",
    booktitle = "Findings of the Association for Computational Linguistics: NAACL 2025",
    month = apr,
    year = "2025",
    address = "Albuquerque, New Mexico",
    publisher = "Association for Computational Linguistics",
    url = "https://aclanthology.org/2025.findings-naacl.379/",
    doi = "10.18653/v1/2025.findings-naacl.379",
    pages = "6814--6821",
    ISBN = "979-8-89176-195-7"
}

@inproceedings{NEURIPS2023_1feb8787,
 author = {Dettmers, Tim and Pagnoni, Artidoro and Holtzman, Ari and Zettlemoyer, Luke},
 booktitle = {Advances in Neural Information Processing Systems},
 doi = {10.52202/075280-0441},
 editor = {A. Oh and T. Naumann and A. Globerson and K. Saenko and M. Hardt and S. Levine},
 pages = {10088--10115},
 publisher = {Curran Associates, Inc.},
 title = {QLoRA: Efficient Finetuning of Quantized LLMs},
 url = {https://proceedings.neurips.cc/paper_files/paper/2023/file/1feb87871436031bdc0f2beaa62a049b-Paper-Conference.pdf},
 volume = {36},
 year = {2023}
}

@inproceedings{NEURIPS2022_8bb0d291,
 author = {Kojima, Takeshi and Gu, Shixiang (Shane) and Reid, Machel and Matsuo, Yutaka and Iwasawa, Yusuke},
 booktitle = {Advances in Neural Information Processing Systems},
 doi = {10.52202/068431-1613},
 editor = {S. Koyejo and S. Mohamed and A. Agarwal and D. Belgrave and K. Cho and A. Oh},
 pages = {22199--22213},
 publisher = {Curran Associates, Inc.},
 title = {Large Language Models are Zero-Shot Reasoners},
 url = {https://proceedings.neurips.cc/paper_files/paper/2022/file/8bb0d291acd4acf06ef112099c16f326-Paper-Conference.pdf},
 volume = {35},
 year = {2022}
}

@article{he2026g2c,
  title={Generate Then Correct: Single-Shot Global Correction for Aspect Sentiment Quad Prediction},
  author={He, Shidong and Wang, Haoyu and Luo, Wenjie},
  journal={arXiv preprint arXiv:2603.13777},
  year={2026},
  url={https://arxiv.org/abs/2603.13777}
}

@inproceedings{wang-etal-2024-context,
    title = "In-Context Example Retrieval from Multi-Perspectives for Few-Shot Aspect-Based Sentiment Analysis",
    author = "Wang, Qianlong  and
      Xu, Hongling  and
      Ding, Keyang  and
      Liang, Bin  and
      Xu, Ruifeng",
    editor = "Calzolari, Nicoletta  and
      Kan, Min-Yen  and
      Hoste, Veronique  and
      Lenci, Alessandro  and
      Sakti, Sakriani  and
      Xue, Nianwen",
    booktitle = "Proceedings of the 2024 Joint International Conference on Computational Linguistics, Language Resources and Evaluation (LREC-COLING 2024)",
    month = may,
    year = "2024",
    address = "Torino, Italia",
    publisher = "ELRA and ICCL",
    url = "https://aclanthology.org/2024.lrec-main.786/",
    pages = "8975--8985"
}

@inproceedings{fan-etal-2025-aspect,
    title = "Aspect-Based Sentiment Analysis with Syntax-Opinion-Sentiment Reasoning Chain",
    author = "Fan, Rui  and
      Li, Shu  and
      He, Tingting  and
      Liu, Yu",
    editor = "Rambow, Owen  and
      Wanner, Leo  and
      Apidianaki, Marianna  and
      Al-Khalifa, Hend  and
      Eugenio, Barbara Di  and
      Schockaert, Steven",
    booktitle = "Proceedings of the 31st International Conference on Computational Linguistics",
    month = jan,
    year = "2025",
    address = "Abu Dhabi, UAE",
    publisher = "Association for Computational Linguistics",
    url = "https://aclanthology.org/2025.coling-main.210/",
    pages = "3123--3137"
}

@ARTICLE{11551863,
  author={Wu, Haiyan and Sun, Chaoqun and Wang, Chenchen and Lu, Chengxiong and Chen, Yanhong and Wang, Jianyong and Zhang, Zhiqiang},
  journal={IEEE Transactions on Knowledge and Data Engineering}, 
  title={TrustSyn: Augmenting LLMs With Constituency- Structured Dependency Knowledge for Aspect-Based Sentiment Analysis}, 
  year={2026},
  volume={38},
  number={8},
  pages={5519-5532},
  doi={10.1109/TKDE.2026.3700652}}

\appendix 
\renewcommand{\thesection}{\Alph{section}}
\section{Related Work}
\label{app:related-work}

\paragraph{Graph-based and Encoder-Level ABSA.}
Early ABSA research commonly incorporated syntactic structure through graph-based neural architectures that propagate sentiment information along dependency relations. Dependency graph convolutional networks \citep{Zhang2019GCN} and attention-enhanced graph variants \citep{HuangCarley2020Attention} improve aspect--opinion interaction modeling by leveraging local syntactic neighborhoods. Large pretrained language models further advanced ABSA performance, with contextual encoders such as BERT substantially outperforming earlier feature-based and recurrent approaches when fine-tuned for downstream sentiment tasks \citep{Devlin2019, Sun2019}. Although these methods benefit from syntactic inductive bias, structural information is typically encoded implicitly within intermediate neural representations, limiting interpretability and making performance sensitive to parsing quality.

\paragraph{Generative and Instruction-Tuned ASQP.}
Recent ASQP methods increasingly formulate quadruple extraction as a unified generative problem. GAS \citep{zhang-etal-2021-towards-generative} casts ASQP as a paraphrase-style sequence generation task, while DLO and ILO \citep{hu-etal-2022-improving-aspect} improve robustness through template-order augmentation. GenDA \citep{wang-etal-2023-generative} further extends this direction with generation-based augmentation, and MvP \citep{gou-etal-2023-mvp} introduces multi-view prompting strategies for improved quadruple generation. IVLS \citep{nie2024modeling} models latent interactions among quadruple elements through implicit variables, whereas MUL \citep{hu-etal-2023-uncertainty} and OTCL \citep{Li2024OpinionTreeguidedCL} enhance robustness through uncertainty-aware learning and contrastive objectives guided by opinion structures.

The emergence of instruction tuning and large language models has further shifted ASQP toward natural-language prompting paradigms \citep{scialom-etal-2022-fine, scaria-etal-2024-instructabsa}. Retrieval-augmented methods additionally improve in-context learning by selecting semantically relevant demonstrations \citep{guu2020realm, zheng-etal-2024-instruction}. 
For few-shot ABSA specifically, \citep{wang-etal-2024-context} show that combining semantic, syntactic, and aspect-sentiment perspectives for demonstration retrieval outperforms similarity-only retrieval, underscoring that retrieval quality itself benefits from structural signal, a complementary observation to DRIFT's focus on injecting structure into the generation prompt rather than the retrieval step.
Despite these advances, most generative and instruction-based approaches still rely primarily on implicit sequential reasoning over surface-form text representations.

\paragraph{Reasoning and Syntax-Aware LLM Methods.}
Several recent studies have explored reasoning-oriented prompting strategies for sentiment analysis. THOR \citep{fei-etal-2023-reasoning} applies chain-of-thought prompting to improve implicit sentiment reasoning, while LLMs for SA \citep{zhang-etal-2024-sentiment} provides a comprehensive evaluation of large language models across sentiment analysis tasks. Syn-Chain \citep{fan-etal-2025-aspect} introduces a zero-shot chain-of-thought framework that sequentially reasons through syntax, opinion extraction, and sentiment classification for general ABSA, distinct from DRIFT's direct linearized-tree injection for the joint ASQP formulation. TrustSyn \citep{11551863} augments LLMs with constituency-structured dependency knowledge for general ABSA, combining phrase-structure and dependency information within a trust-augmented representation, distinct from DRIFT's direct linearization of dependency-only trees for the joint ASQP formulation. More recently, SimRP \citep{Jian_Chen_Li_Wang_Zeng_Yao_An_Wu_2025} incorporates syntactic similarity into retrieval-based demonstration selection for in-context learning. However, existing approaches typically use syntax either internally within encoder architectures or externally for retrieval selection, rather than explicitly exposing dependency structures as relational reasoning guidance during generation itself.

\paragraph{Recent ASQP Directions and Evaluation.}
Recent work has explored alternative ASQP formulations and robustness strategies. Self-training with pseudo-label scoring \citep{zhang-etal-2024-self-training} improves generative robustness through iterative refinement, while black-box LLM alignment methods \citep{li-etal-2025-aligning-black} investigate instruction-based adaptation without direct parameter access. Structured prediction approaches such as grid tagging \citep{su-etal-2025-unified} reformulate ASQP as token-level relational labeling, whereas pseudo-token and review-combination methods \citep{chen-etal-2025-using} improve representation efficiency and contextual aggregation. Stepwise syntax integration tuning \citep{chen-etal-2025-s2it} incrementally incorporates syntactic cues during model adaptation, but does not inject complete dependency structures directly into prompts as explicit reasoning context.

Beyond extraction itself, recent studies have shown that LLMs can function as semantic evaluators capable of assessing equivalence between generated outputs and reference annotations. However, ASQP evaluation still relies predominantly on exact-match metrics that are sensitive to span-boundary variation and paraphrastic differences. In contrast to prior approaches, DRIFT explicitly linearizes full dependency trees and injects them into both generation and evaluation prompts, enabling structure-grounded reasoning and syntax-aware semantic adjudication within a unified LLM-based ASQP framework.

We follow the reporting protocol of SimRP to ensure fair comparison across retrieval-based settings.

\section{Implementation Setup}
\label{app:set-up}
\subsection{Datasets}
\label{app:datasets}
We perform experiments on four ASQP benchmarks spanning two 
domains. For the restaurant domain, we use Rest15 
(SemEval-2015 Task~12) \citep{Pontiki2015} and Rest16 
(SemEval-2016 Task~5) \citep{Pontiki2016}, as well as 
Restaurant-ACOS \citep{Cai2021ACOS}. For the laptop domain, 
we include Laptop-ACOS \citep{Cai2021ACOS}. Each sentence 
is labeled with one or more quadruples \((a, c, s, o)\) 
denoting the \emph{aspect term}, \emph{category}, 
\emph{sentiment polarity}, and \emph{opinion term}.

The SemEval datasets (Rest15 and Rest16) follow standard 
train/validation/test splits and ASQP preprocessing 
conventions. The ACOS datasets provide train/dev/test splits following 
the version used in \citet{Cai2021ACOS}. For ICL experiments on all four datasets, no fine-tuning 
data is used; the training split serves solely as the 
demonstration retrieval pool. Dataset statistics are provided in 
Tables~\ref{tab:dataset-stats} and~\ref{tab:acos-stats}.

\begin{table}[t]
\centering
\small
\resizebox{\columnwidth}{!}{%
\begin{tabular}{l ccc ccc}
\toprule
\multirow{2}{*}{\textbf{Dataset}} &
\multicolumn{3}{c}{\textbf{Rest15}} &
\multicolumn{3}{c}{\textbf{Rest16}} \\
\cmidrule(lr){2-4}\cmidrule(lr){5-7}
& Train & Valid & Test & Train & Valid & Test \\
\midrule
N & 834 & 209 & 537 & 1264 & 316 & 544 \\
\midrule
Positive & 1005 & 252 & 453 & 1369 & 341 & 583 \\
Neutral  & 34 & 14 & 37 & 62 & 23 & 40 \\
Negative & 315 & 81 & 305 & 558 & 143 & 176 \\
\midrule
\textbf{Total} & 1354 & 347 & 795 & 1989 & 507 & 799 \\
\bottomrule
\end{tabular}%
}
\caption{Dataset statistics for Rest15 and Rest16. N denotes the number of sentences. 
Positive, Neutral, and Negative represent the number of quadruples for each sentiment.}
\label{tab:dataset-stats}
\end{table}

\begin{table}[t]
\centering
\small
\resizebox{\columnwidth}{!}{%
\begin{tabular}{l ccc ccc}
\toprule
\multirow{2}{*}{\textbf{Dataset}} &
\multicolumn{3}{c}{\textbf{Rest-ACOS}} &
\multicolumn{3}{c}{\textbf{Laptop-ACOS}} \\
\cmidrule(lr){2-4}\cmidrule(lr){5-7}
& Train & Dev & Test & Train & Dev & Test \\
\midrule
N & 1530 & 171 & 583 & 2934 & 326 & 816 \\
\midrule
Positive & 1656 & 180 & 667 & 2583 & 279 & 716 \\
Neutral & 95 & 12 & 44 & 227 & 24 & 65 \\
Negative & 733 & 69 & 205 & 1362 & 137 & 380 \\
\midrule
\textbf{Total} & 2484 & 261 & 916 & 4172 & 440 & 1161 \\
\bottomrule
\end{tabular}%
}
\caption{Dataset statistics for Restaurant-ACOS and Laptop-ACOS.
N denotes the number of sentences. Positive, Neutral, and Negative
represent the number of quadruples for each sentiment.}
\label{tab:acos-stats}
\end{table}

\subsection{Evaluation Metrics}
\label{app:evaluation-metrics}
We adopt F1 as the primary metric and also report precision and recall. A prediction is considered correct only when all four elements of a quadruple \emph{aspect term}, \emph{category}, \emph{polarity}, and \emph{opinion term} exactly match the corresponding gold labels. Evaluation uses normalized spans and exact matching across all four elements.

\subsection{Implementation Details}
\label{app:implementation-details}

We fine-tune Meta-Llama-3-8B-Instruct\footnote{\href{https://huggingface.co/meta-llama/Meta-Llama-3-8B-Instruct}{\textnormal{https://huggingface.co/meta-llama/Meta-Llama-3-8B-Instruct}}} using QLoRA with 4-bit NF4 quantization and train with LoRA (Low-Rank Adaptation) \citep{lora2022hu} techniques with a single A100-40GB GPU. Training employs the paged AdamW optimizer with a learning rate of $2\times10^{-4}$, cosine learning rate decay, and a warmup ratio of $0.05$ over $4$ epochs. The effective batch size is $16$ achieved by accumulating gradients $4$ times with a batch size of $4$. For LoRA, we set the rank $r=128$, resulting in approximately 500 million trainable parameters. This configuration allows us to capture high-dimensional task-specific information while training only a small fraction of the total model weights, significantly reducing the VRAM overhead compared to full fine-tuning. We choose Llama-3-8B-Instruct for its strong instruction-following ability and efficient trade-off between performance and compute, ideal for parameter-efficient fine-tuning.

\section{Multi-Seed Results}
\label{app:seeds}

To assess the stability of the main PEFT result reported in
Section~\ref{sec:main-results}, we repeated training and evaluation across
5 random seeds, with all hyperparameters (Appendix~\ref{app:implementation-details})
held fixed and only the random seed controlling model initialization,
data shuffling, and sampling varied between runs. Table~\ref{tab:seeds}
reports the per-seed F1 scores on Rest15 and Rest16, alongside the mean
and standard deviation.

Across both datasets, the standard deviation remains below 0.15 F1
(0.13 on Rest15, 0.08 on Rest16), indicating that performance is stable
across initializations and that the headline result does not depend on a
favorable random seed. The headline result reported in
Section~\ref{sec:main-results} (54.47/64.10) corresponds to the best
single run among these five seeds, consistent with the single-run
reporting convention adopted by comparison methods including SimRP. We
report the full multi-seed distribution here, rather than only the best
run, to allow direct comparison of run-to-run variance against future
work.

\begin{table}[t]
\centering
\small
\begin{tabular}{lcc}
\toprule
\textbf{Seed} & \textbf{Rest15 F1} & \textbf{Rest16 F1} \\
\midrule
1        & 54.20 & 64.10 \\
2        & 54.13 & 64.05 \\
3        & 54.33 & 63.89 \\
4        & 54.47 & 63.95 \\
5        & 54.28 & 64.03 \\
\midrule
\textbf{Mean $\pm$ Std} & \textbf{54.28 $\pm$ 0.13} & \textbf{64.00 $\pm$ 0.08} \\
\bottomrule
\end{tabular}
\caption{Per-seed F1 (\%) for the PEFT configuration with dependency-tree injection (no retrieval), across 5 random seeds.}
\label{tab:seeds}
\end{table}

\section{Prompt Specifications and Dependency Linearization}
\label{app:prompts}
We provide the full prompt specifications and dependency linearization 
format below to ensure reproducibility.

\subsection{Dependency Tree Linearization}
\label{app:linearization}
Dependency parses are generated using spaCy. For each token $t$ in a sentence, we construct the representation:

\[
token.text - token.head.text \rightarrow token.dep
\]
For example, given the sentence:
\begin{quote}
\emph{``The food was absolutely delicious.''}
\end{quote}
the extracted dependency triples include:
\begin{align*}
\text{food} &\rightarrow \text{nsubj} \rightarrow \text{was}; \\
\text{delicious} &\rightarrow \text{acomp} \rightarrow \text{was}; \\
\text{absolutely} &\rightarrow \text{advmod} \rightarrow \text{delicious}.
\end{align*}
The linearized representation injected into the prompt is:
\begin{quote}
\texttt{[DEP] food$\rightarrow$nsubj$\rightarrow$was;}\\
\texttt{delicious$\rightarrow$acomp$\rightarrow$was;}\\
\texttt{absolutely$\rightarrow$advmod$\rightarrow$delicious [/DEP]}
\end{quote}
All tokens are serialized into a newline-separated list. This linearized representation is inserted into the prompt to guide aspect–opinion linking.

\subsection{PEFT Prompt}

For PEFT-based instruction tuning with Llama-3 8B, we use the following system instruction:

\begin{small}
\begin{lstlisting}
Task: Extract aspect-sentiment quadruples from the query: [aspect, category, sentiment, opinion_word].
\end{lstlisting}
\end{small}

The input sentence and dependency tree are appended to this system message to form the full training prompt.

\subsection{ICL Prompt for GPT-4o / GPT-4o-mini}

\begin{small}
\begin{lstlisting}
ROLE:
You are an expert ASQP (aspect sentiment quadruple prediction) annotation bot. Your processing must be precise, systematic, and strictly follow all rules.

TASK DEFINITION:
Your sole task is to extract all sentiment quadruples from a user-provided sentence.
Your response MUST be a valid Python list of lists, with no other text or explanations.
The output format for each quadruple is a list of four strings:
[<Aspect Term>, <Aspect Category>, <Sentiment Polarity>, <Opinion Term>]

ELEMENT DEFINITIONS:
1.  Aspect Term: The specific noun or phrase being reviewed.
    - It MUST be a term from the restaurant domain (e.g., "food", "service", "waiter").
    - Use the exact string "NULL" if the opinion is general or the aspect is implicit.
2.  Aspect Category: The general class for the Aspect Term.

3.  Sentiment: The polarity of the opinion.
    - It MUST be one of three strings: "positive", "negative", or "neutral".
4.  Opinion Term: The exact word(s) from the sentence expressing the sentiment.

EXTRACTION RULES:
1.  Concise Spans: The Opinion Term must be the most concise phrase that still captures the full sentiment. Do not include unnecessary words.
2.  Extract All Opinions: You must extract a separate quadruple for every distinct opinion expressed in the sentence.
3.  Implied Sentiment: Analyze comparative and conditional phrases to determine the implied sentiment.

STEP BY STEP PROCESS:
To ensure accuracy, think step-by-step before generating the final list:
1.  First, identify all opinion-expressing words/phrases in the sentence.
2.  Second, for each opinion, identify its specific aspect phrase. If none exist, that aspect is "NULL".
3.  Third, determine the sentiment polarity (positive, negative, neutral) and the correct aspect category from the provided list.
4.  Finally, construct the list of all quadruples according to all rules above.

FINAL OUTPUT INSTRUCTION:
Your final output MUST ONLY be a valid Python list of lists.
- Do not include explanations, reasoning, or markdown formatting.
- Do not include extra text before or after the list.
- Each inner list must contain exactly 4 strings.
\end{lstlisting}
\end{small}
\medskip
\textit{Note: For Laptop-ACOS experiments, the domain 
description is adjusted to the laptop domain 
(e.g., ``battery'', ``screen'', ``keyboard'').}
\medskip

\subsection{ICL Prompt for Llama-3.1}
\begin{small}
\begin{lstlisting}
ROLE:
You are an expert ASQP (aspect sentiment quadruple prediction) annotation bot. Your processing must be precise, systematic, and strictly follow all rules.

TASK DEFINITION:
Your sole task is to extract all sentiment quadruples from a user-provided sentence and its dependency tree.
Your response MUST be a valid Python list of lists, with no other text or explanations.
The output format for each quadruple is a list of four strings:
[<Aspect Term>, <Aspect Category>, <Sentiment Polarity>, <Opinion Term>]

ELEMENT DEFINITIONS:
1.  Aspect Term: The specific noun or phrase being reviewed.
    - It MUST be a term from the restaurant domain (e.g., "food", "service", "waiter").
    - Use the exact string "NULL" if the opinion is general or the aspect is implicit.
2.  Aspect Category: The general class for the Aspect Term.

3.  Sentiment: The polarity of the opinion.
    - It MUST be one of three strings: "positive", "negative", or "neutral".
4.  Opinion Term: The exact word(s) from the sentence expressing the sentiment.

EXTRACTION RULES:
1.  Concise Spans: The Opinion Term must be the most concise phrase that still captures the full sentiment. Do not include unnecessary words.
2.  Extract All Opinions: You must extract a separate quadruple for every distinct opinion expressed in the sentence.
3.  Dependency Tree Usage: To ensure accuracy, you must use the dependency tree to find the correct relationship between an opinion and an aspect. Follow the grammatical connections in the tree to link an Opinion Term back to the specific Aspect Term it modifies. This is the most reliable way to create a correct quadruple. If no such link exists, the Aspect Term is "NULL".
4.  Implied Sentiment: Analyze comparative and conditional phrases to determine the implied sentiment.

STEP BY STEP PROCESS:
To ensure accuracy, think step-by-step before generating the final list:
1.  First, identify all opinion-expressing words/phrases in the sentence.
2.  Second, for each opinion, identify its specific aspect phrase. If none exist, that aspect is "NULL".
3.  Third, determine the sentiment polarity (positive, negative, neutral) and the correct aspect category from the provided list.
4.  Finally, construct the list of all quadruples according to all rules above.

FINAL OUTPUT INSTRUCTION:
Your final output MUST ONLY be a valid Python list of lists.
- Do not include explanations, reasoning, or markdown formatting.
- Do not include extra text before or after the list.
- Each inner list must contain exactly 4 strings.
- Use double quotes for strings, not single quotes.
\end{lstlisting}
\end{small}

\subsection{Example Prompt Construction}

Input sentence:

\begin{small}
\begin{lstlisting}
The food was delicious but the service was slow.
\end{lstlisting}
\end{small}

Dependency tree:

\begin{small}
\begin{lstlisting}
food - was -> nsubj
delicious - was -> acomp
service - was -> nsubj
slow - was -> acomp
but - delicious -> cc
\end{lstlisting}
\end{small}

Constructed prompt:

\begin{small}
\begin{lstlisting}
Sentence:
The food was delicious but the service was slow.

Dependency Tree:
food - was -> nsubj
delicious - was -> acomp
service - was -> nsubj
slow - was -> acomp
but - delicious -> cc
\end{lstlisting}
\end{small}

\subsection{LLM-as-a-Judge Prompt}

\begin{small}
\begin{lstlisting}
ROLE: Expert Linguistic Analyst and Judge

### OBJECTIVE:
You must evaluate which of two candidate extractions of aspect-sentiment quadruples is more correct, based strictly on the rules below.

### TASK DEFINITION
A valid quadruple = [Aspect, Category, Sentiment, Opinion Word]

**RULES:**
1. All elements must be grounded in the query sentence.
2. Opinion Word must appear verbatim in the sentence.
3. Aspect & Category must be logically consistent. 
4. Sentiment must accurately reflect the tone of the Opinion Word.

### CONTEXT:
- Query Sentence: {sentence}
- Dependency Tree: {dependency_tree}

### OPTIONS TO JUDGE:
Option 0: {opt0}
Option 1: {opt1}

### ASSIGNMENT:
Determine which option is more accurate according to the rules.

Return only JSON with keys "reasoning" and "decision".
- reasoning: explanation for your judgment
- decision: 0 if Option 0 is correct, 1 if Option 1 is correct
\end{lstlisting}
\end{small}

\section{Qualitative Examples}
\label{app:qualitative}

Table~\ref{tab:qualitative-examples} presents five representative cases in
which exact-match evaluation marks a DRIFT prediction as incorrect, while
we judge the prediction to be semantically equivalent to, or
more accurate than, the reference annotation. We discuss each case below,
organized according to the four sources of boundary-level disagreement
introduced in Section~\ref{sec:span-boundary}: (i) additional opinion terms
captured by the model but absent from the reference, (ii) incorrect aspect
attachment in the reference annotation, (iii) implicit aspect or category
grounding, and
(iv) flexible opinion span boundaries.

\paragraph{Additional Opinion Terms.}
In \emph{``Great place to relax and enjoy your dinner,''} the reference
contains a single quadruple, [place, ambience general, positive, Great],
capturing only the adjective \textit{Great} as the opinion expression. DRIFT
additionally proposes [place, ambience general, positive, enjoy], correctly
recognizing that the verb \textit{enjoy} also carries positive sentiment
toward the same aspect \textit{place}. Because the sentence expresses two
syntactically distinct but semantically compatible opinion cues attached to
the same aspect, exact-match evaluation penalizes DRIFT for the additional,
valid quadruple that the reference happens to omit. We agreed
that both extractions are correct, indicating that annotation completeness,
rather than model error, is the source of the mismatch.

\paragraph{Incorrect Aspect Attachment in the Reference.}
Two examples illustrate cases where the reference annotation itself
misattaches an opinion term to the wrong aspect. In \emph{``The hot dogs are
good, yes, but the reason to get over here is the fantastic pork croquette
sandwich, perfect on its supermarket squishy bun,''} the reference incorrectly
assigns the opinion \textit{perfect} to the aspect \textit{bun}, whereas the
syntactic and semantic context indicates that \textit{perfect} modifies the
\textit{pork croquette sandwich} as a whole, with \textit{bun} only a
compositional part of it. DRIFT's dependency-guided prompt correctly
identifies \textit{sandwich} as the syntactic target of \textit{perfect},
omitting the spurious \textit{bun} quadruple present in the reference. A
similar attachment ambiguity occurs in \emph{``I've had my fair share of
modern Japanese, and this spot delivers,''} where the reference attaches
\textit{delivers} to \textit{modern Japanese} (the cuisine), while DRIFT
correctly attaches it to \textit{spot} (the restaurant), consistent with the
dependency path linking \textit{delivers} to its syntactic subject
\textit{this spot} rather than to the object of the preceding clause. In both
cases, the dependency structure exposed by DRIFT provides an explicit signal
for resolving attachment ambiguity that exact-match comparison alone cannot
capture.

\paragraph{Implicit Aspect/Category Grounding.}
The remaining examples concern category or aspect assignment under implicit
or general opinions. In \emph{``Which lets face it \ldots\ at times
it's a good thing,''} both the reference and DRIFT correctly leave the aspect
term as \textit{NULL}, since the sentence names no explicit target. However,
the reference assigns the category \textit{service general}, which has no
lexical anchor anywhere in the sentence, whereas DRIFT assigns the broader
and better-grounded category \textit{restaurant miscellaneous}, correctly
reflecting that the sentence expresses a general, non-aspect-specific
positive sentiment without invoking service at all. Conversely, in \emph{``In the end you
end up with a fair tab and NOTHING BUT A GREAT TIME!!!,''} DRIFT identifies
the explicit aspect term \textit{tab}, supported by the dependency structure
linking \textit{fair} directly to \textit{tab}, which the reference omits in
favor of \textit{NULL} -- indicating that the reference under-specifies an
aspect that is, in fact, explicitly present in the surface form.

\paragraph{Flexible Opinion Span Boundaries.}
The same \textit{tab}/\textit{GREAT TIME} example also illustrates
span-boundary flexibility: the reference marks the opinion span as
\textit{GREAT}, while DRIFT extracts the longer span \textit{GREAT TIME}.
Both spans express equivalent positive sentiment toward the same implicit
aspect (\textit{restaurant general}), and we confirmed that the
longer span is, if anything, more informative without altering the underlying
polarity. This case exemplifies the central motivation for our LLM-based
adjudication mechanism: minor differences in span boundary do not necessarily
indicate a genuine extraction error.

Across all five cases, the disagreements arise from annotation granularity, attachment ambiguity, and span-boundary convention rather than model error. DRIFT consistently predicts complete, well-grounded aspect–opinion pairs even under ambiguous or implicit conditions, supporting the case for syntax-aware semantic adjudication as a necessary complement to rigid exact-match scoring.

\FloatBarrier
\begin{table*}[t!]
\vspace*{-18cm}
\centering
\scriptsize
\setlength{\tabcolsep}{3pt}
\renewcommand{\arraystretch}{1.2}
\begin{tabularx}{\textwidth}{
  >{\raggedright\arraybackslash}p{0.17\textwidth}
  >{\raggedright\arraybackslash}p{0.19\textwidth}
  >{\raggedright\arraybackslash}p{0.19\textwidth}
  c
  c
  >{\raggedright\arraybackslash}X}
\toprule
\textbf{Sentence (Query)} & \textbf{Reference} & \textbf{Prediction} &
\textbf{Exact Match} & \textbf{Human} & \textbf{Reasoning} \\
\midrule

Great place to relax and enjoy your dinner.
& {[place, ambience general, positive, Great]}
& {[place, ambience general, positive, Great]}
  {[place, ambience general, positive, enjoy]}
& \xmark & \cmark
& Includes both \textit{Great} and \textit{enjoy}; reference omits one. \\
\addlinespace[2pt]

The hot dogs are good, yes, but the reason to get over here is the
fantastic pork croquette sandwich, perfect on its supermarket squishy bun.
& {[hot dogs, food quality, positive, good]}
  {[pork croquette sandwich, food quality, positive, fantastic]}
  {[bun, food quality, positive, perfect]}
& {[hot dogs, food quality, positive, good]}
  {[pork croquette sandwich, food quality, positive, fantastic]}
& \xmark & \cmark
& Reference wrongly assigns \textit{perfect} to \textit{bun} instead of
the sandwich. \\
\addlinespace[2pt]

In the end you end up with a fair tab and NOTHING BUT A GREAT TIME!!!
& {[NULL, restaurant prices, positive, fair]}
  {[NULL, restaurant general, positive, GREAT]}
& {[tab, restaurant prices, positive, fair]}
  {[NULL, restaurant general, positive, GREAT TIME]}
& \xmark & \cmark
& Correctly identifies \textit{tab} as aspect; reference misses it. \\
\addlinespace[2pt]

Which lets face it \ldots\ at times it's a good thing.
& {[NULL, service general, positive, good]}
& {[NULL, restaurant miscellaneous, positive, good]}
& \xmark & \cmark
& Reference assigns an ungrounded \textit{service general} category; sentence never mentions service. \\
\addlinespace[2pt]

I've had my fair share of modern Japanese, and this spot delivers.
& {[modern Japanese, food quality, positive, delivers]}
& {[spot, restaurant general, positive, delivers]}
& \xmark & \cmark
& \textit{delivers} refers to \textit{spot}, not \textit{modern Japanese}. \\

\bottomrule
\end{tabularx}
\caption{Qualitative examples where model predictions are semantically correct despite failing exact-match evaluation.
\cmark\ = correct, \xmark\ = incorrect.}
\label{tab:qualitative-examples}
\end{table*}

\end{document}